\documentclass[10pt,twocolumn,letterpaper]{article}
\usepackage[pagenumbers]{cvpr} 

\usepackage{float}
\usepackage{graphicx} 
\usepackage{natbib}
\usepackage{amsmath}
\usepackage{amssymb}
\usepackage{amsfonts}
\usepackage{mathtools}
\usepackage[table,xcdraw,dvipsnames]{xcolor}

\usepackage{tikz}
\usetikzlibrary{arrows.meta}

\usepackage{adjustbox}
\usepackage{enumitem}
\usepackage{subcaption}
\usepackage{ifthen}
\usepackage{wrapfig}
\usepackage[most]{tcolorbox}
\tcbuselibrary{skins}
\usepackage{makecell}
\definecolor{cvprblue}{rgb}{0.21,0.49,0.74}
\definecolor{boxblue}{HTML}{507494}
\definecolor{boxgray}{HTML}{8a8a8a}
\definecolor{boxred}{HTML}{805353}
\definecolor{boxorange}{HTML}{bd9b77}

\makeatletter
\let\save@mathaccent\mathaccent
\newcommand*\if@single[3]{%
  \setbox0\hbox{${\mathaccent"0362{#1}}^H$}%
  \setbox2\hbox{${\mathaccent"0362{\kern0pt#1}}^H$}%
  \ifdim\ht0=\ht2 #3\else #2\fi
  }
\newcommand*\rel@kern[1]{\kern#1\dimexpr\macc@kerna}
\newcommand*\widebar[1]{\@ifnextchar^{{\wide@bar{#1}{0}}}{\wide@bar{#1}{1}}}
\newcommand*\wide@bar[2]{\if@single{#1}{\wide@bar@{#1}{#2}{1}}{\wide@bar@{#1}{#2}{2}}}
\newcommand*\wide@bar@[3]{%
  \begingroup
  \def\mathaccent##1##2{%
    \let\mathaccent\save@mathaccent
    \if#32 \let\macc@nucleus\first@char \fi
    \setbox\z@\hbox{$\macc@style{\macc@nucleus}_{}$}%
    \setbox\tw@\hbox{$\macc@style{\macc@nucleus}{}_{}$}%
    \dimen@\wd\tw@
    \advance\dimen@-\wd\z@
    \divide\dimen@ 3
    \@tempdima\wd\tw@
    \advance\@tempdima-\scriptspace
    \divide\@tempdima 10
    \advance\dimen@-\@tempdima
    \ifdim\dimen@>\z@ \dimen@0pt\fi
    \rel@kern{0.6}\kern-\dimen@
    \if#31
      \overline{\rel@kern{-0.6}\kern\dimen@\macc@nucleus\rel@kern{0.4}\kern\dimen@}%
      \advance\dimen@0.4\dimexpr\macc@kerna
      \let\final@kern#2%
      \ifdim\dimen@<\z@ \let\final@kern1\fi
      \if\final@kern1 \kern-\dimen@\fi
    \else
      \overline{\rel@kern{-0.6}\kern\dimen@#1}%
    \fi
  }%
  \macc@depth\@ne
  \let\math@bgroup\@empty \let\math@egroup\macc@set@skewchar
  \mathsurround\z@ \frozen@everymath{\mathgroup\macc@group\relax}%
  \macc@set@skewchar\relax
  \let\mathaccentV\macc@nested@a
  \if#31
    \macc@nested@a\relax111{#1}%
  \else
    \def\gobble@till@marker##1\endmarker{}%
    \futurelet\first@char\gobble@till@marker#1\endmarker
    \ifcat\noexpand\first@char A\else
      \def\first@char{}%
    \fi
    \macc@nested@a\relax111{\first@char}%
  \fi
  \endgroup
}
\makeatother

\definecolor{citeblue}{HTML}{2d576b}
\definecolor{darkgreen}{HTML}{61b06e}
\definecolor{lightgreen}{HTML}{eef6ef}
\definecolor{darkred}{HTML}{bf5f50}
\definecolor{trainable-orange}{HTML}{fccb9f}
\definecolor{pointblue}{HTML}{156082}
\definecolor{pointgreen}{HTML}{4ea72e}
\definecolor{flowgreen}{HTML}{4ea72e}
\definecolor{flowblue}{HTML}{0f9ed5}

\makeatletter
\newcommand{\captionLink}[2]{%
  {
    \renewcommand{\@makecaption}[2]{%
        \vskip\abovecaptionskip
        \sbox\@tempboxa{\textcolor{citeblue}{\href{#1}{\underline{##1}}}: ##2}%
        \ifdim \wd\@tempboxa >\hsize
            \textcolor{citeblue}{\href{#1}{\underline{##1}}}: ##2\par
        \else
            \global \@minipagefalse
            \hb@xt@\hsize{\hfil\box\@tempboxa\hfil}%
        \fi
        \vskip\belowcaptionskip}%
    \caption{#2}%
  }
}
\makeatother

\newcommand\norm[1]{\left\lVert#1\right\rVert}

\newcommand{\R}{\mathbb R}

\newcommand{\E}{\mathbb E}

\newcommand{\myparagraph}[1]{\vspace{-2.5mm}\paragraph*{#1}}
\newcommand{\mysubsection}[1]{\subsection{#1}\vspace{-1.5mm}}
\newcommand{\mysection}[1]{\section{#1}\vspace{-1mm}}

\usepackage[pagebackref,breaklinks,colorlinks,citecolor=cvprblue]{hyperref}

\title{Gaussians-to-Life: Text-Driven Animation of 3D Gaussian Splatting Scenes}

\author{Thomas Wimmer\textsuperscript{1,2} \quad Michael Oechsle\textsuperscript{3} \quad Michael Niemeyer\textsuperscript{3} \quad Federico Tombari\textsuperscript{1,3}\\
{\textsuperscript{1}Technical University of Munich \quad
\textsuperscript{2}Max Planck Institute for Informatics \quad
\textsuperscript{3}Google}\\
{\small\url{https://wimmerth.github.io/gaussians2life.html}} 
}

\begin{document}
\twocolumn[{
\maketitle
\vspace{-2\intextsep}
\begin{center}
    \captionsetup{type=figure}
    \centering
    \includegraphics[width=\textwidth]{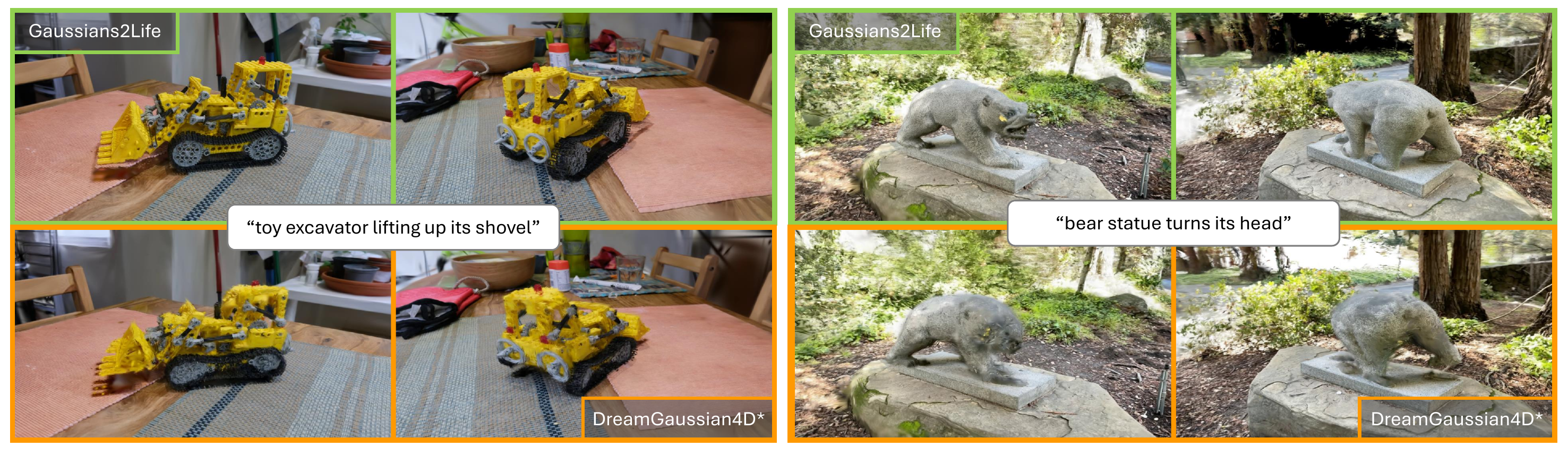}\vspace{-.7\intextsep}
    \captionof{figure}{Our proposed method Gaussians2Life preserves high visual quality of scenes while animating them according to a text prompt. It significantly outperforms a baseline method crafted from DreamGaussian4D~\citep{ren2023dreamgaussian4d} and creates more realistic movements.}
    \label{fig:teaser-figure}
\end{center}
}]

\begin{abstract}
State-of-the-art novel view synthesis methods achieve impressive results for multi-view captures of static 3D scenes. However, the reconstructed scenes still lack “liveliness,” a key component for creating engaging 3D experiences. Recently, novel video diffusion models generate realistic videos with complex motion and enable animations of 2D images, however they cannot naively be used to animate 3D scenes as they lack multi-view consistency. To breathe life into the static world, we propose \emph{Gaussians2Life}, a method for animating parts of high-quality 3D scenes in a Gaussian Splatting representation. Our key idea is to leverage powerful video diffusion models as the generative component of our model and to combine these with a robust technique to lift 2D videos into meaningful 3D motion. We find that, in contrast to prior work, this enables realistic animations of complex, pre-existing 3D scenes and further enables the animation of a large variety of object classes, while related work is mostly focused on prior-based character animation, or single 3D objects. Our model enables the creation of consistent, immersive 3D experiences for arbitrary scenes.
\end{abstract}

\mysection{Introduction}

Recent advances in 3D representation, like NeRF and 3D Gaussian Splatting (3DGS)~\citep{mildenhall2021nerf,kerbl20233d}, have emerged as powerful tools for highly accurate and fast novel view rendering of static scenes. While these approaches enable immersive experiences with impressive quality, their static nature can result in a lack of dynamism and engagement. Breathing life into scenes and automatically generating animation of previously static 3D objects holds strong potential to create more engaging and realistic experiences.

In the 2D domain, video diffusion models have demonstrated significant capabilities in generating realistic animations in videos given input images or text prompts~\citep{ho2022imagen, singer2022make, blattmann2023align, blattmann2023stable}.
However, recent advances to leverage such models for the creation of dynamic 3D content~\citep{singer2023text,bahmani20234d,ling2023align,ren2023dreamgaussian4d,yu20244real, ren2024l4gm} still lag behind the generative capabilities of video diffusion models, particularly animating existing 3D scenes appears to be underexplored and methods are limited to animate single assets~\citep{jiang2024animate3d, ren2023dreamgaussian4d}. To this end, we investigate how realistic outputs from existing video diffusion models can help animating objects in static 3D scenes. More specifically, we aim to animate 3DGS scenes following a user-defined text prompt and a bounding box containing the target object.
We identified two key challenges to solve this task. The first challenge is about how to generate multi-view consistent video guidance with a VDM for a static scene. Given valid video guidance, the remaining challenge is how to lift generated 2D videos into realistic and consistent 3D motions of the 3DGS primitives in the static scene without degrading the visual quality of the input scene.

In this work, we provide a method capable of addressing these challenges. By leveraging multi-view information in the video diffusion step, we can generate approximately multi-view consistent video guidance without the need for expensive fine-tuning of the diffusion model. Further, we provide an in-depth analysis of lifting 2D motion to 3D and propose a robust framework for animating 3DGS scenes by using video diffusion guidance in a pipeline that combines depth estimation and point tracking to generate 3D anchor trajectories. These are used to animate the static 3DGS scene in a multi-view consistent manner.
In summary, our contributions are as follows:
\begin{itemize}
    \item We introduce a novel method for animating 3DGS scenes given a text prompt and an object bounding box. 
    \item Our approach interfaces existing open-source video diffusion models to generate multi-view consistent video guidance for a static scene and lifts the 2D video guidance to a realistic 3D motion for Gaussian primitives in the 3DGS scene.
    \item We provide an experimental evaluation on real-world scenes from the MipNeRF360~\citep{barron2022mip} and Instruct-NeRF2NeRF~\citep{haque2023instructnerf} datasets, where we utilize an adaptation of DreamGaussian4D~\citep{ren2023dreamgaussian4d} as a relevant baseline. Further, we provide an in-depth ablation study indicating the effectiveness of our architectural choices.
\end{itemize}

\mysection{Related Work}
\vspace{1mm}
\myparagraph{Text-to-Video Generation}
The recent success of text-to-image diffusion models~\citep{ho2020denoising, rombach2022high} has increased interest in generative models for other data types, including videos. Common paradigms in the creation of video diffusion models are to build upon pre-trained 2D image generative models and to train additional components for modeling temporal relationships between generated video frames~\citep{ho2022imagen, singer2022make, blattmann2023align, blattmann2023stable}.
Recent works have also proposed additional conditioning signals besides text, e.g.,~images or sparse manually defined motion~\citep{blattmann2023stable, wang2023modelscope, xing2023dynamicrafter, wang2023videocomposer}.
We employ video generative models in our approach to optimize dynamics in a given 3D scene. The resulting 4D scenes are naturally 3D-consistent and can be rendered in real-time from any viewpoint, a major advantage over 4D generative models in many use cases.
Recent, concurrent works explored explicit camera control or multi-view generation for video diffusion models~\citep{xie2024sv4d, he2024cameractrl, kuang2024collaborative, bahmani2024vd3d}. We note that multi-view consistent outputs are not guaranteed for these models, as they are not based on an explicit 3D representation. Additionally, rendering novel views of the dynamic scene requires querying the diffusion model, which is costly and cannot be performed in real-time.

\myparagraph{Dynamic Gaussian Splatting}
\citet{kerbl20233d} proposed 3D Gaussian Splatting (3DGS), a method for novel view synthesis that bridges the gap between volumetric-rendering based implicit representations~\citep{mildenhall2021nerf} and explicit 3D representations, like 3D meshes, by representing a 3D scene as a set of 3D Gaussians.
The explicit nature of 3DGS allows for fast rasterization and more controllability in modeling.

Naturally, follow-up works also explored the reconstruction of dynamic 3D scenes~\citep{luiten2023dynamic, yang2023deformable, wu20234d, huang2023sc, lin2023gaussian, das2023neural}. The most common strategy in this domain is the optimization of a canonical 3DGS representation alongside a neural deformation field that maps a 3D coordinate $x$ and time $t$ to attribute changes at the respective moment $t$ for the 3D Gaussians at $x$ in the canonical representation. Various regularization terms have been proposed to steer the optimized motion to be physically or geometrically plausible, i.e., to preserve local rigidity~\citep{luiten2023dynamic, huang2023sc}, isometry~\citep{luiten2023dynamic}, or momentum~\citep{duisterhof2023md}. \citet{gao2024gaussianflow} further proposed using an off-the-shelf 2D flow estimation model as additional supervision signal for deformations. Recent concurrent works proposed methods for monocular dynamic scene reconstruction that make use of different pre-trained 2D models to obtain more information on the 3D structure~\citep{stearns2024dynamic, lei2024mosca}.
While we also make use of pre-trained 2D models to lift motion into 3D, we generate dynamics instead of reconstructing them. For this, we repeatedly generate new guidance videos and lift motion from different viewpoints.

\myparagraph{4D Generative Models} \label{sec:lit-4d-generation}
\citet{poole2022dreamfusion} proposed Score Distillation Sampling (SDS) to leverage 2D diffusion models as powerful priors for 3D generative tasks, which was followed by several technical improvements~\citep{wang2023score, wang2024prolificdreamer, yu2023text}. In SDS, 3D scenes are rendered at every optimization step, 2D views are noised, and one de-noising step with the 2D diffusion model is performed, which provides a gradient signal that can be used for back-propagation to the parameterized scene. An alternative to SDS is multi-step denoising~\citep{zhou2023sparsefusion, wu2023reconfusion}, which follows the same idea as SDS, but instead of doing one, multiple de-noising steps, as well as decoding from latent to pixel space are performed, similarly to the standard inference of a diffusion model. Losses are subsequently computed directly in pixel space.

Recently, several works started exploring the use of SDS for 4D generation~\citep{singer2023text,bahmani20234d,ling2023align,ren2023dreamgaussian4d,yu20244real, ren2024l4gm}. While some methods directly perform text-to-4d generation, others require image or video inputs which first need to be generated using off-the-shelf 2D diffusion models. \citet{bahmani2024tc4d} proposed trajectory-conditioned 4D generation, where coarse trajectories of objects are given as an additional input, which resolves the problem of limited motion in other 4D generation works.
Except for 4Real~\citep{yu20244real}, a concurrent work, all methods are restricted to single animated 3D objects, often lacking photo-realism. The main reason for this focus on object-centric generation is the usage of multi-view image diffusion models that are trained on datasets of single 3D objects and do not generalize to more complex 3D scenes~\citep{liu2023zero,shi2023mvdream}. Instead of generating dynamic 3DGS scenes from only text, our goal is to generate dynamics for a given 3DGS scene and text as control.
Another concurrent work, Animate3D~\citep{jiang2024animate3d}, is the only work that we are aware of that aims at animating given 3D scenes. Contrary to our work, this method focuses on the animation of single 3D assets, while we aim at generating realistic deformations in the contexts of larger 3D scenes.
Finally, another recent direction for animation of 3D scenes is the optimization of physical material fields~\citep{zhang2024physdreamer, huang2024dreamphysics}, which can then be used to perform physically realistic animations based on manual inputs and the PhysGaussian method~\citep{xie2023physgaussian}. We note that such methods usually require manipulation by external agents or external forces, e.g.,~using deformation handles, and are limited to a handful of simulation types~\citep{xie2023physgaussian}. In contrast, our proposed method is able to synthesize any type of deformation using only text as a control signal.

\mysection{Method}
A successful method for animating 3D scenes requires two main components: A powerful driving signal for motion generation and an effective way to distill this motion into the 3D scene while keeping the scene appearance and generating motion realistic. The diffusion model guidance should stay closely aligned with the given 3D scene, as well as be as multi-view consistent as possible, while the distillation of this signal into the 3D world should be as efficient as possible.  scenario, these two components will improve each other to achieve the best possible results.
Our method offers improvements over standard SDS- and optimization-based solutions by introducing a training-free approach for generating approximately multi-view consistent outputs of a video diffusion model and a technique to directly lift 2D motion into 3D, leveraging several pre-trained 2D models to align information between generated videos and the given 3D scene.

\mysubsection{Basic Setup}
We formalize our problem as follows: We are given a captured 3DGS scene as input, as well as a text prompt that describes the desired motion within the scene.
Each Gaussian in the initial scene has a mean position $\mu \in \R^3$, an anisotropic covariance matrix, factorized into a scaling vector $s \in \R^3$ and a quaternion rotation $q \in \R^4$, as well as an opacity level $\alpha$ and view-dependent colors represented using spherical harmonics. In our work, we aim to add a temporal dimension to the given 3D scene. While we do not change the opacity of Gaussians, the position $\mu_t$, scaling $s_t$, and rotation $q_t$ should be time-dependent attributes, see Sec.~\ref{sec:2d-to-3d-lifting}.

For more user control, we allow for a user-defined selection of scene elements that should be animated. Such selection can stem from binary labeling of 3D Gaussians, which can be automated using open-world 3D segmentation methods~\citep{qin2023langsplat} and is often assumed given in related works~\citep{luiten2023dynamic, zhang2024physdreamer}, or from simple 3D bounding boxes.

Our method is agnostic to the specific 3DGS implementation used for capturing as long as the output follows the standard 3DGS conventions.
During optimization, we only use the 0-degree spherical harmonics to save computational resources but note that at inference time, higher-degree spherical harmonics can be added back again while being rotated similarly as the corresponding 3D Gaussians, see \citet{xie2023physgaussian}.

\mysubsection{Diffusion Guidance}
In this section, we describe how we interface a recent video diffusion model for the purpose of generating dynamics for a given 3DGS scene.

\myparagraph{Image-Conditioned Generation}
The use of an image-conditioned diffusion model~\citep{brooks2023instructpix2pix} has proven to be beneficial for 3D scene editing~\citep{haque2023instructnerf}. As outputs are more aligned with the given 3D scenes through such conditioning, the noise level in SDS can be increased, resulting in larger amounts of motion compared to outputs of solely text-conditioned diffusion guidance, where the noise level needs to be reduced to stay aligned with the 3D scene. We thus employ a text- and image-conditioned video diffusion model, DynamiCrafter~\citep{xing2023dynamicrafter}, as guidance in our method.

\myparagraph{Multi-Step Denoising}
Score Distillation Sampling suffers from several technical issues, as pointed out by previous works~\citep{wang2023score, wang2024prolificdreamer, yu2023text}. Proposed solutions often require fine-tuning or training a second diffusion model, which is impractical for video diffusion models. Besides that, SDS imposes a computational burden where the diffusion model must be queried at each optimization step, and the loss computed in latent space needs to be back-propagated to the scene parameterization.

Instead, we propose using multi-step denoising (\cref{sec:lit-4d-generation}), which decouples generation from optimization, as pixel-level outputs (videos) can be stored and reused. This approach also enables the computation of additional supervision signals, like optical flow or depth, which is not possible with SDS. Moreover, pixel-level outputs improve user control during optimization, addressing the instability of current text-to-4D methods in public video diffusion models.

\begin{figure*}[ht]
    \centering
    \includegraphics[width=\textwidth]{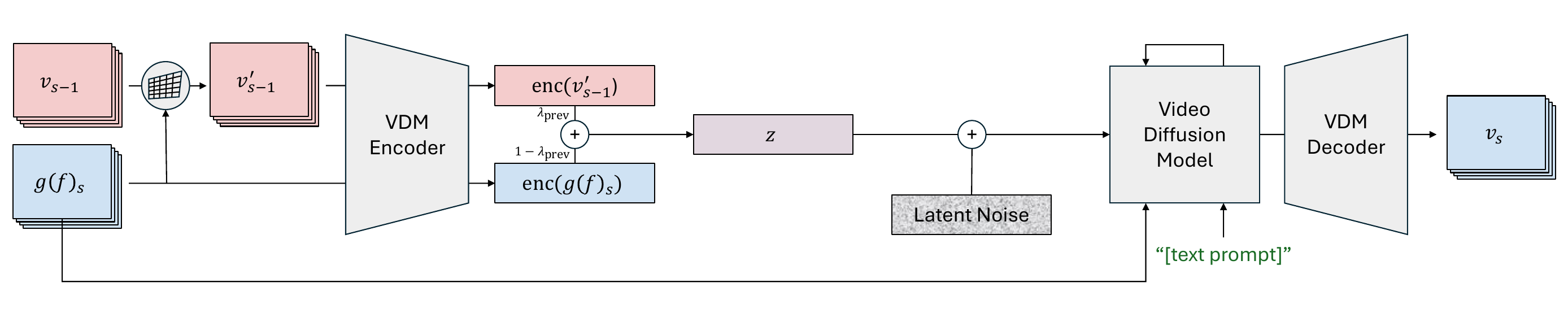}\vspace{-.5\intextsep}
    \caption{Improvement of multi-view consistency of generated videos through latent interpolation. In addition to the rendering of the dynamic scene $f$ using the rendering function $g$ from the current viewpoint $g(f)_{s}$, we compute the latent embedding of the warped video output $v_{s-1}$ of the previous optimization step $s-1$ (from a different viewpoint). We linearly interpolate the latents before passing them through the video diffusion model (VDM), which is additionally conditioned on the static scene view from the current viewpoint. The resulting output is finally decoded to a new video output $v_{s}$.}
    \label{fig:mv-diffusion}
    \vspace{-1\intextsep}
\end{figure*}

\myparagraph{Multi-View Consistent Video Generation}
A remaining problem of current video diffusion models is their lack of output consistency. Especially when generating videos from different viewpoints, generated motion in the videos will often be inconsistent and thus hinder the optimization of 3D dynamics. For static 3D objects, fine-tuning image diffusion models for generating multi-view outputs is a standard solution for improving this multi-view consistency. This fine-tuning, however, is even more costly and difficult with video diffusion models, where multi-view data is limited and fine-tuning on synthetic datasets, e.g., of single animated 3D objects, limits the generalizability of the video diffusion models~\citep{xie2024sv4d}. 

Instead, we leverage the 3D information given through the static scene initialization. While using rendered views from the current viewpoint promotes 3D awareness of the diffusion model, this signal is only static. To steer generations to also contain consistent motion, we propose a new method of latent interpolation, where we encode the previously generated guidance video $v_{s-1}$ and fuse its latent, which encodes the desired motion, with the latent of the current video rendering $g(f)_s$, as shown in \cref{fig:mv-diffusion}:
\begin{equation}
    z = \lambda_\textrm{prev}\, \textrm{enc}(v_{s-1}) + (1-\lambda_\textrm{prev})\, \textrm{enc}(g(f)_s),
\end{equation}
where $z$ denotes the fused latent, $g(\cdot)$ the rendering function, $f$ the 4D Gaussian Splatting scene, and $\lambda_\textrm{prev}$ is a hyperparameter that is gradually decreased with increasing number of steps $s$. 
To resemble motion from the new viewpoint as realistically as possible, we make use of an off-the-shelf optical flow estimation model to warp the video frames of $v_{s-1}$ as detailed in \cref{app:flow-warping}. While this warping can, of course, not give us a truly realistic projection of the motion to the current viewpoint, it helps in adapting the video to resemble the view from the new viewpoint when using small viewpoint changes. We adapt the view sampling procedure to reflect this assumption of small baseline changes, as described in \cref{app:viewpoint-sampling}.

We note that this proposed method puts a stronger emphasis on the initial diffusion video that is created for the first viewpoint. This fits well for the case where the user selects this first guidance video but can potentially cause problems when the pipeline is run without such a selection and the first generated video does not represent realistic motion. In such cases, the next generated outputs can still make up for any previous mistakes, but there is no guarantee of improvements. We include an exemplary multi-view generation of our proposed method in \cref{fig:mv-diffusion-example}.

\begin{figure}[ht]
    \centering
    \includegraphics[width=\columnwidth]{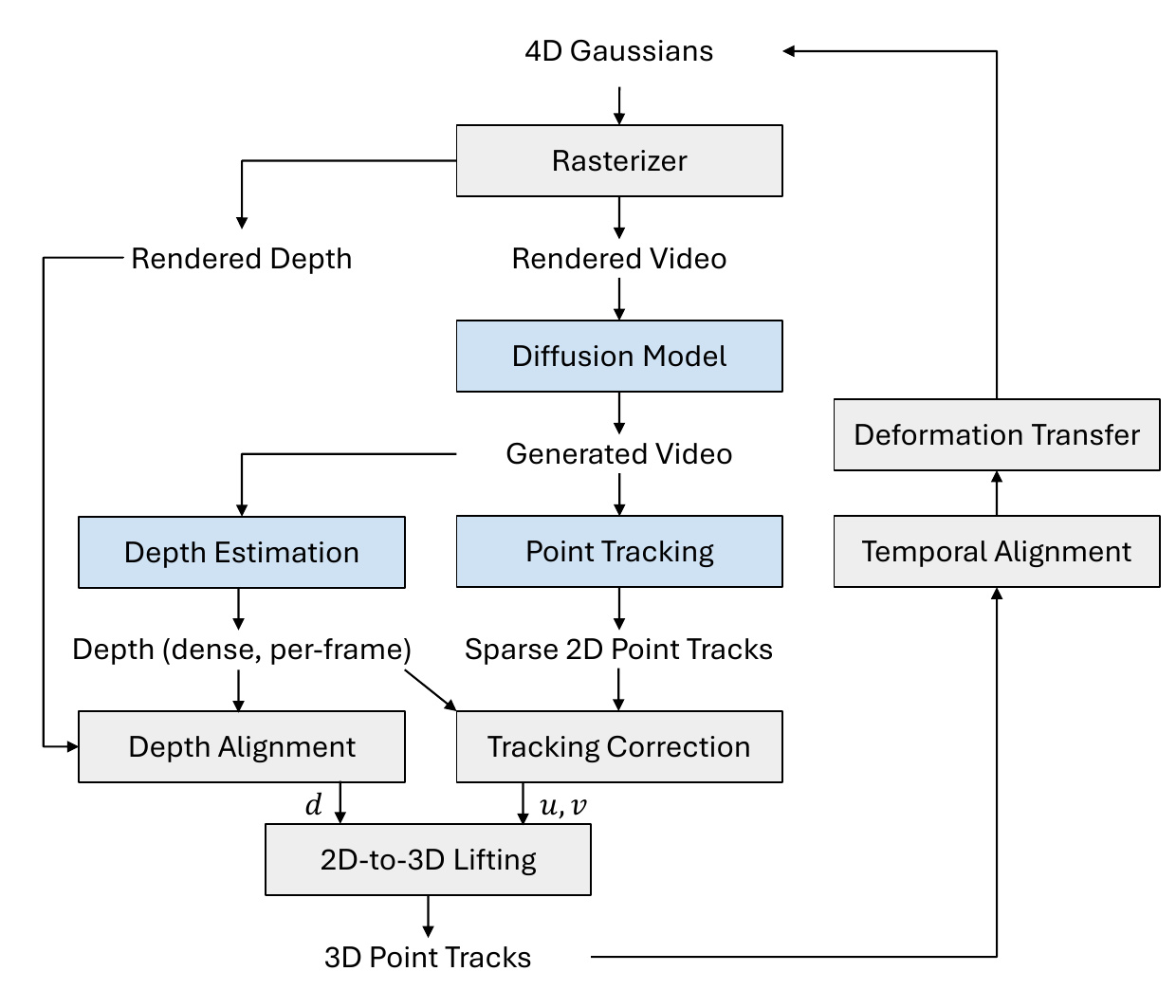}\vspace{-.5\intextsep}
    \caption{Pipeline for lifting 2D dynamics into 3D. Pre-trained models are shown in \textcolor{boxblue}{blue}. We detect 2D point tracks and use aligned estimated depth values to lift them into 3D.}
    \label{fig:trajectory-lifting}
    \vspace{-1\intextsep}
\end{figure}

\mysubsection{Lifting 2D Motion to 3D} \label{sec:2d-to-3d-lifting}
Given valid video guidance, we now investigate the second core question: how can we lift 2D guidance signals efficiently to 3D? We first point out that optimization-based solutions, i.e., SDS or rendering-based optimizations, are slow in convergence and sensitive to the still-existing small inconsistencies in the generated guidance videos. Results are, therefore, often noisy and do either yield divergent motion or almost static outputs (see Ablation studies, \cref{fig:ablations}). To evade these problems, we propose to instead leverage the power of several pre-trained 2D models to lift motion from 2D to 3D.
More specifically, we combine 2D point tracking and depth estimation to obtain information on the depicted 3D motion from the videos, similar to concurrent works for monocular dynamic reconstruction~\citep{lei2024mosca, stearns2024dynamic}. In our case, however, we use the 2D model outputs to efficiently bring dynamics into the 3D scene in just one step, compared to multiple steps necessary in optimization-based solutions~\citep{ling2023align, ren2023dreamgaussian4d}). We repeat this procedure from other viewpoints to get more reliable and more 3D-aware estimations. We explain our method that is schematically visualized in \cref{fig:trajectory-lifting} in the following.

We propose to track a sparse set of points throughout the generated video that is sampled from the video frame depicting the static scene rendering\footnote{The video diffusion model used in our method~\citep{xing2023dynamicrafter} was trained to always contain the image condition at one frame in the video output, which is often but not necessarily the first frame.} (which we refer to as $t_0$). Recent works~\citep{karaev2023cotracker} have made remarkable progress in this task, being able to reliably track a set of points and even giving accurate estimations and recover points that are occluded. Further, we use an off-the-shelf metric depth estimation model~\citep{piccinelli2024unidepth} to obtain dense per-pixel depth estimations for every frame.

\myparagraph{Tracking Correction}
Using 2D point tracks $(u,v)_t$ and per-frame dense depth maps, we compute the depth $d_t$ for all tracked points at each timestep. Despite the robustness of the point-tracking method, errors can occur when the tracker ``loses'' a point and begins tracking another in the background or foreground. To address this, we use a correction method that detects errors by thresholding the depth value ratio between consecutive frames, where the threshold value is chosen manually to be $\max\{d_t, d_{t+1}\} / \min\{d_t, d_{t+1}\} < 1.2$ in our experiments. 
If the relative difference is large, we assume a tracking error and correct it by checking the local pixel neighborhood for a better tracking point, i.e., with a smaller depth ratio. If none is found, we discard the respective trajectory.
We estimate depth values for points at frames where they are not visible with a cubic spline interpolation from known depth values. When extrapolation is needed, we use linear extrapolation based on the interpolation gradients at the last visible point.

\myparagraph{Depth Alignment}
For every tracked point $p_i$, we compare the estimated depth $d_i$ at the video frame at $t_0$ with the ground-truth depth value $d^{\textrm{ GT}}_i$ for the given scene and compute the ratio between them, which we then apply to the respective estimated depth values in other frames:
\begin{equation}
    d'_{i,t} = d_{i,t} \frac{d_{i,t_0}}{d^{\textrm{ GT}}_i}.
\end{equation}
We note that as we sample the tracking points at the same frame, all tracked points are always visible in it.

\myparagraph{2D-to-3D Lifting}
As camera extrinsics $R, T$ and intrinsics $K$ of the rasterization camera are known, we can project point trajectories given by pixel coordinates $(u,v)_t$ and estimated depth $d_t$ back into the 3D world space $X_t$:

\begin{equation}
    X_t = \begin{bmatrix}
R \
T \\
0 \
1
\end{bmatrix}^{-1}
\begin{bmatrix}
    K^{-1}
    \begin{bmatrix}
        u_t\
        v_t\
        d_t
    \end{bmatrix}^T \
    1
\end{bmatrix}^T
\end{equation}

To reduce memory and computational demands, we only store the trajectories that lie within the 3D bounding boxes of the animated scene elements at $t_0$.

\myparagraph{Temporal Alignment}
Knowing the timestep $t_0$, where the video shows the static scene, allows us to temporally align trajectories at this point, as shown in \cref{fig:temporal-alignment}. During optimization, we use a rendered video as noised input to the diffusion model, which is fixed at generating $n$ frames. As we wish to avoid interpolation, we select the sequence with the most overlaps to include maximum information from sampled viewpoints. If multiple sequences have the same ``support,'' we select the last one, containing $t_0$ as early as possible. During testing, we can make use of linear interpolation between discrete timesteps to extend the frame count in the generated scene renderings.

\begin{figure}[ht]
    \centering
    \includegraphics[width=.7\columnwidth]{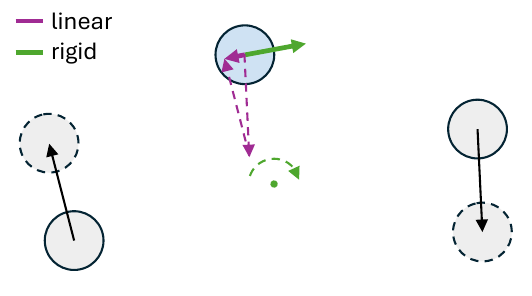}\vspace{-.5\intextsep}
    \caption{Comparison of linear and rigid motion estimation. The rigid motion estimation finds a fitting rotation for the \textcolor{boxgray}{source displacements} and estimates the displacement for the \textcolor{boxblue}{target point} accordingly.}
    \label{fig:linear-vs-rigid-motion-estimation}
    \vspace{-.5\intextsep}
\end{figure}

\begin{figure*}[htb!]
    \centering
    \begin{subfigure}[c]{\textwidth}
        \centering
        \includegraphics[width=.9\textwidth]{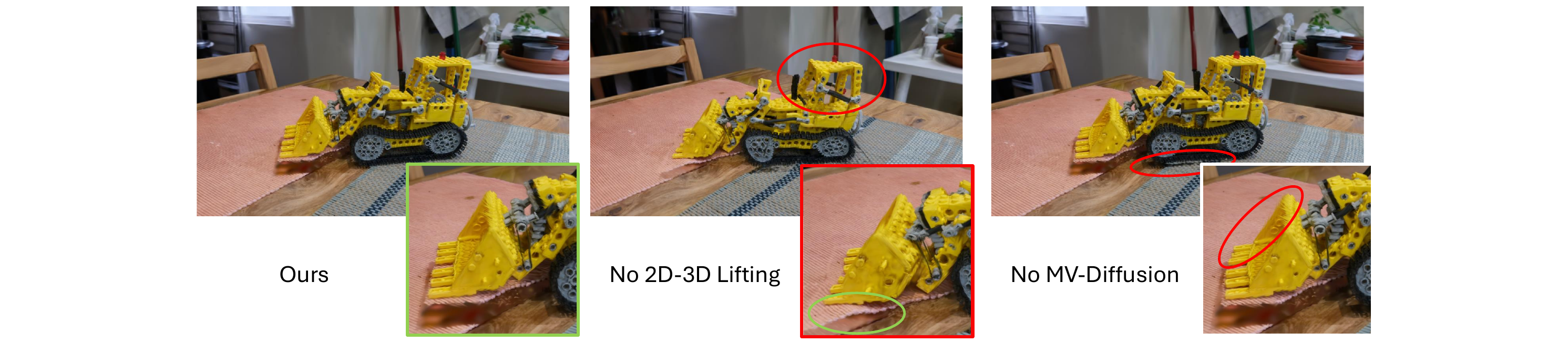}
        \caption{Comparison against ablated versions.}
    \end{subfigure}

    \vspace{.5\intextsep}
    \begin{subfigure}[c]{\textwidth}
        \centering
        \includegraphics[width=.9\textwidth]{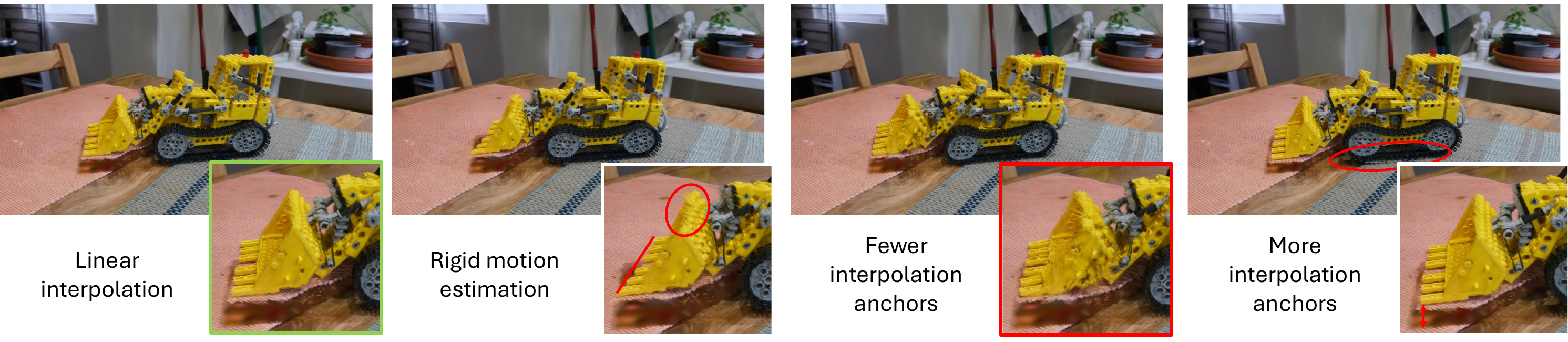}
        \caption{Motion transfer from anchor trajectories to 3D Gaussians.}
    \end{subfigure}\vspace{-.5\intextsep}
    \caption{Qualitative comparison against ablations on the LEGO bulldozer scene for the prompt ``toy bulldozer lifting its shovel.''}
    \label{fig:ablations}
    \vspace{-.5\intextsep}
\end{figure*}

\myparagraph{Deformation Transfer} \label{sec:motion-transfer}
Given the projected point trajectories, referred to as anchor trajectories in the following, we still need to transfer the motion onto single 3D Gaussians. At this point, the explicit nature of 3DGS comes in handy, as we can use deformation estimation methods that are inspired by techniques from traditional geometry processing (see \cref{fig:linear-vs-rigid-motion-estimation}). As we can directly infer motion that is, e.g., as rigid as possible while closely aligned with the anchor trajectories, this also is a decisive advantage over optimization-based solutions using such terms as regularization, where such terms hinder the amount of motion being distilled, as no motion is the most rigid motion possible.

Firstly, we propose to use a weighted linear motion estimation (cf. Linear Blend Skinning).
To compute the weights, first, the $K$ nearest anchor trajectories at $t_0$ and their distance $d$ to the center of the Gaussian are determined for each 3D Gaussian. The displacement is then calculated as a weighted average of the neighboring anchor trajectory displacements $y_j-x_j$, where the weight depends on the distance $d$:
\begin{align} \label{eq:method-linear-interpolation}
    t_i &= \sum_{j \in k\textrm{-NN}(i, K)} w_{i,j} (y_j - x_j), \\
    \textrm{with}\quad w_{i,j} &= \cfrac{\exp(-\tau d_{i,j})}{\sum_{j' \in k\textrm{-NN}(i, K)} \exp(-\tau d_{i,j})}
\end{align}
and a temperature parameter $\tau$. To estimate rotation and scaling changes of 3D Gaussians, one can use a similar method as the subsequently proposed rigid motion estimation with fixed $t$.

A second, more sophisticated technique is the estimation of rigid body movements from the computed displacements of the anchor points. To do so, we make use of the Kabsch algorithm~\citep{kabsch1976solution, umeyama1991least} to estimate the optimal rotation, isotropic scaling and translation that align the point clouds of neighboring anchor trajectories at two subsequent timesteps, where we choose the weights $w_{i,j}$ as in Eq.~\ref{eq:method-linear-interpolation}:
\begin{equation}
    \min_{R_i,t_i,s_i} \sum_{j \in k\textrm{-NN}(i, K)} w_{i,j} \norm{s_iR_ix_j+t_i-y_j}^2.
\end{equation}

We note that though always considering the $K$ nearest neighbors for every 3D Gaussian, motion is being refined when adding information from more guidance videos from different viewpoints.

\mysection{Experiments}
In this section, we provide experimental results of our method and qualitatively compare it to an adapted version of DreamGaussian4D~\citep{ren2023dreamgaussian4d}. Further, we demonstrate the effectiveness of our methodological choices in an in-depth ablation study. In our experiments, we focus on qualitative analysis due to the lack of established metrics for generative 3D dynamics (without ground truth deformations), as well as the lack of competing methods. However, we provide a section on possible metrics (and their shortcomings) as well as quantitative results for our ablation study in \cref{sec:quantitative-ablation}.

\mysubsection{Setup}
\vspace{1.5mm}
\vspace{5pt}
\myparagraph{Dataset}
We select a number of scenes from the Mip-NeRF 360~\citep{barron2022mip} dataset, where we use RadSplat~\citep{niemeyer2024radsplat} for 3D reconstruction, as well as the bear scene from the Instruct-NeRF2NeRF~\citep{haque2023instructnerf} dataset, which we reconstruct with standard Gaussian Splatting~\citep{kerbl20233d}.

\myparagraph{Baseline}
Given that no prior work tackled the exact task of animating objects in the context of a full 3D scene, we carefully compose a baseline by adapting DreamGaussian4D~\citep{ren2023dreamgaussian4d} for the considered setting\footnote{The concurrent method Animate3D~\citep{jiang2024animate3d} for text-driven animation of single objects has no public code at the time of submission.}.
As this method was developed for single-object video-to-4D generation, we can also use it to demonstrate the strengths and importance of animating objects within a larger scene context. DreamGaussian4D works by first creating a static 3DGS model that is subsequently deformed following a given 2D video, as well as using SDS with a multi-view diffusion model~\citep{liu2023zero}. For a fair comparison, we use the bounding boxes or masks that our method takes as input to mask out background elements for \citep{ren2023dreamgaussian4d}. Further, we use the same initial diffusion guidance videos, where the background is automatically removed for the input video in DreamGaussian4D. We provide more details on the baseline, as well as the used hyperparameters in \cref{sec:supp-baseline-dreamgaussian4d,sec:supp-baseline-hyperparameters}.

\mysubsection{Qualitative Results}
We show qualitative results\footnote{We strongly recommend the reader to check the \href{https://wimmerth.github.io/gaussians2life.html}{project website} for the videos corresponding to the respective figures.} of our method applied to the different scenes along with the corresponding text prompts in \cref{fig:qualitative-results}. As can be seen, our method is able to generate compelling deformation while preserving the visual quality of the initial 3D scenes. We additionally show the optical flow between different frames, where we employ the same color coding as \citet{baker2011database}.

\mysubsection{Baseline Comparison}
We show exemplary comparisons of our method against the DreamGaussian4D baseline on real-world scenes in \cref{fig:teaser-figure}. We notice two things: First, the visual quality of the 3D scene is much more preserved by our method. This can be attributed to several factors. Our method is based on deforming the existing objects by inferring 3D motion directly from generated videos. The detected anchor trajectories are then used to estimate the transformation of the single 3D Gaussians. As the nearest neighboring anchor trajectories for 3D Gaussians that lie close to each other have significant overlap, shapes are automatically smoothly deformed. On the opposite, the baseline deforms the 3D Gaussians independently from each other, resulting in less coherent motion. The appearance-based optimization is also more prone to artifacts in the diffusion model guidance and, as the employed diffusion model is not image-conditioned, can also exhibit problems like the Janus problem which is not explicitly solved by the multi-view diffusion model, see the bear from the back side in \cref{fig:teaser-figure}.

Next, we notice that our method is able to generate more realistic movements of the animated objects \textit{within} the context of the larger scenes. As diffusion guidance is generated for the objects within the initial, larger scene instead of for only the single object, contact points and realistic motion within the scene are implicitly taken care of. The baseline that uses a multi-view diffusion model that was trained on single 3D assets is not able to model the motion within this scene context, leading to discontinuities with the background elements, e.g., the feet of the bear in \cref{fig:teaser-figure}.

\mysubsection{Ablation studies} \label{sec:qualitative-ablation}
In our ablation study, we first perform a comparison against an optimization-based approach, a commonly used technique for modeling dynamic scenes where a neural field is trained to model the deformations of the scene~\citep{ren2023dreamgaussian4d, ling2023align}, guided by our proposed diffusion guidance. More details on this baseline model are provided in \cref{sec:supp-baseline-ours}. To show the effects of our proposed approximately multi-view consistent diffusion, we further ablate a version of our method using standard novel-view video generation without the proposed latent interpolation. Finally, we compare rigid and linear motion estimation and analyze the effect of the number of anchor trajectories considered in motion transfer.
The qualitative results of this analysis can be found in \cref{fig:ablations}.

While our method effectively deforms 3D scenes by lifting 2D motion to 3D and transferring it to 3D Gaussians, rendering-based optimization of a deformation field fails to generate coherent and realistic motion. The baseline particularly struggles with the temporal consistency of the generated motion due to the lack of robustness of rendering-based optimizations to inconsistent input data (see, e.g., static 3D Gaussian Splatting reconstruction of moving objects as in \citet{wu20234d}, Fig.~5). Although our proposed approximately multi-view consistent video generation mitigates this issue to some extent, the remaining inconsistencies still pose a significant challenge for optimizing meaningful motion. On the other hand, appearance-based optimization is able to ``fill'' holes in the 3D scene by moving or scaling Gaussians accordingly, which is an advantage over our purely deformation-based approach.
The use of our proposed approximately multi-view diffusion helps increase 3D consistency in the results and especially reduces diffusion artifacts, which can lead to noisy anchor trajectories.

\setlength{\intextsep}{0pt}%
\setlength{\columnsep}{4pt}%
\begin{wrapfigure}{R}{.5\columnwidth}
    \centering
    \vspace{-3mm}
    \includegraphics[width=.4\columnwidth]{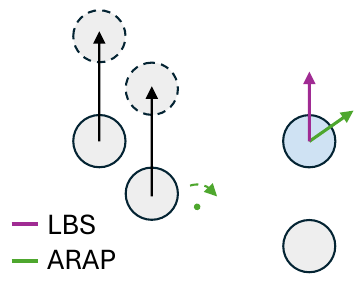}
    \caption{Linear vs. rigid motion estimation with limited observations.}
    \label{fig:lbs-vs-arap-failure}
\end{wrapfigure}

\begin{figure*}[htb!]
    \centering
    \setlength{\intextsep}{12pt}%
    \vspace{-.5\intextsep}
    \setlength{\intextsep}{0pt}%
    \includegraphics[width=\textwidth]{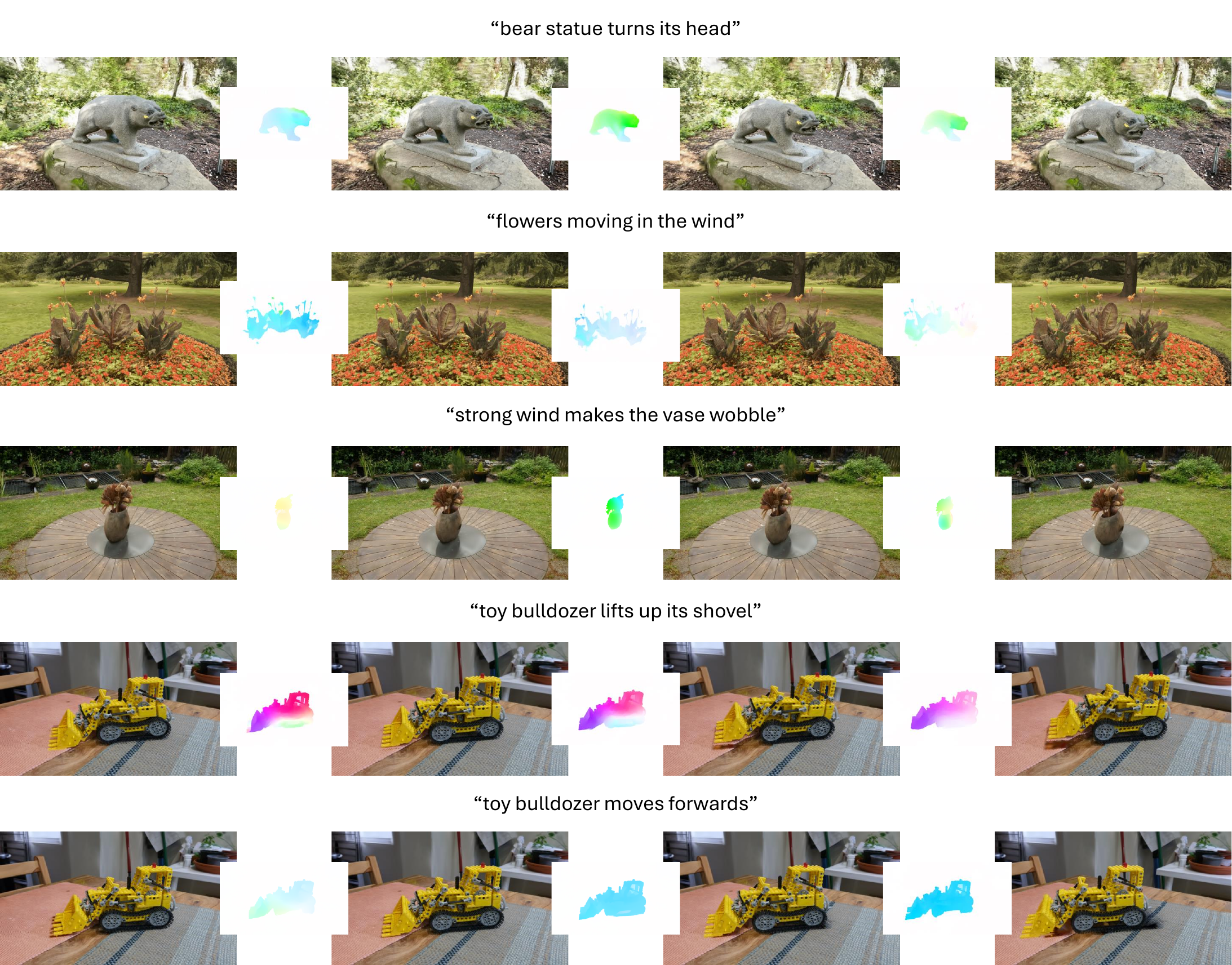}
    \caption{Qualitative results of our method on different 3D scenes with the optical flow (see \cref{fig:flow-cmap}) shown between keyframes.}
    \label{fig:qualitative-results}
    \vspace{-2\intextsep}
\end{figure*}
When transferring motion from anchor trajectories to 3D Gaussians, linear motion estimation surprisingly often results in more rigid deformation than rigid motion estimation, as seen at the back ends of the shovel in \cref{fig:ablations}. The rigid motion estimation struggles with unbalanced observations, e.g., in the first optimization step. Specifically, when most registered (visible) anchor trajectories are on one side of the object, while some static observed points lie behind the Gaussians in question, as simplified in Fig.~\ref{fig:lbs-vs-arap-failure}, where the query point (blue) is influenced by two moving source points and one static point, the rigid motion estimation predicts an incorrect rotation of the object. Subsequent optimization steps from other viewpoints fail to correct this error as noise is gradually removed from scene renderings, causing the diffusion model to adapt to the faulty movement.

\mysection{Conclusions}
We presented Gaussians2Life, a technique for text-driven animation of static 3D Gaussian Splatting scenes based on video diffusion models. We addressed shortcomings of current public video diffusion models to improve 3D consistency in the outputs and enhance motion distillation from 2D into static 3D scenes while keeping the initial scene appearance intact through geometry-aware deformation transfer techniques. We emphasize that our proposed method is optimization-free and thus significantly faster than a baseline using an inference-time optimization. The high-quality results of applying our method to diverse real-world captures underline its strength in the faithful animation of given 3D scenes.
\vspace{-2mm}
\myparagraph{Limitations} \label{sec:main-limitations}
The proposed method deforms an existing scene without adding or removing 3D Gaussians. Thus, it is not possible to fill holes created by moving objects (see \cref{fig:ablations}) or to add and remove particles, e.g., fire, in an animation.
As our method does not include any rendering-based optimization, eventual artifacts through moving Gaussians are not made up for. A possible fix to this problem is employing SDS to refine the generated motion in an appended stage, e.g., as proposed by \citet{bahmani2024tc4d}.
Object intersections or collisions are not handled explicitly but implicitly by using a generalist video diffusion model that generates videos of objects moving in the full scenes. While this generally works, making up for wrong depth or tracking estimations is not possible.

Finally, we note that there exists a domain mismatch between the often inherently static 3D scenes captured using 3D reconstruction methods and the dynamic scenes used in the training of video diffusion models. While this mismatch is limiting the effectiveness of diffusion model guidance, we introduced several techniques to deal with noisy predictions in this work. Generally, we observe that with currently available ``open'' video diffusion models, often, a realistic \textit{actor} is still necessary for animating scenes, which can also be invisible, such as ``the wind.''
\vspace{-2mm}
\myparagraph{Acknowledgements}
We thank Daniel Cremers, whose computing resources were used for some of the experiments in the context of this work.
TW is supported by the Konrad Zuse School ELIZA, sponsored by the German Federal Ministry of Education and Research.

\setlength{\intextsep}{12pt}%
\setlength{\columnsep}{22pt}%

{
    \small
    \bibliographystyle{ieeenat_fullname}
    \bibliography{main}

\begin{thebibliography}{63}
\providecommand{\natexlab}[1]{#1}
\providecommand{\url}[1]{\texttt{#1}}
\expandafter\ifx\csname urlstyle\endcsname\relax
  \providecommand{\doi}[1]{doi: #1}\else
  \providecommand{\doi}{doi: \begingroup \urlstyle{rm}\Url}\fi

\bibitem[Abdal et~al.(2024)Abdal, Yifan, Shi, Xu, Po, Kuang, Chen, Yeung, and Wetzstein]{abdal2023gaussian}
Rameen Abdal, Wang Yifan, Zifan Shi, Yinghao Xu, Ryan Po, Zhengfei Kuang, Qifeng Chen, Dit-Yan Yeung, and Gordon Wetzstein.
\newblock Gaussian shell maps for efficient 3d human generation.
\newblock In \emph{2024 IEEE/CVF Conference on Computer Vision and Pattern Recognition (CVPR)}, page 9441–9451. IEEE, 2024.

\bibitem[Bahmani et~al.(2024{\natexlab{a}})Bahmani, Liu, Yifan, Skorokhodov, Rong, Liu, Liu, Park, Tulyakov, Wetzstein, Tagliasacchi, and Lindell]{bahmani2024tc4d}
Sherwin Bahmani, Xian Liu, Wang Yifan, Ivan Skorokhodov, Victor Rong, Ziwei Liu, Xihui Liu, Jeong~Joon Park, Sergey Tulyakov, Gordon Wetzstein, Andrea Tagliasacchi, and David~B. Lindell.
\newblock \emph{TC4D: Trajectory-Conditioned Text-to-4D Generation}, page 53–72.
\newblock Springer Nature Switzerland, 2024{\natexlab{a}}.

\bibitem[Bahmani et~al.(2024{\natexlab{b}})Bahmani, Skorokhodov, Rong, Wetzstein, Guibas, Wonka, Tulyakov, Park, Tagliasacchi, and Lindell]{bahmani20234d}
Sherwin Bahmani, Ivan Skorokhodov, Victor Rong, Gordon Wetzstein, Leonidas Guibas, Peter Wonka, Sergey Tulyakov, Jeong~Joon Park, Andrea Tagliasacchi, and David~B. Lindell.
\newblock 4d-fy: Text-to-4d generation using hybrid score distillation sampling.
\newblock In \emph{2024 IEEE/CVF Conference on Computer Vision and Pattern Recognition (CVPR)}, page 7996–8006. IEEE, 2024{\natexlab{b}}.

\bibitem[Bahmani et~al.(2024{\natexlab{c}})Bahmani, Skorokhodov, Siarohin, Menapace, Qian, Vasilkovsky, Lee, Wang, Zou, Tagliasacchi, et~al.]{bahmani2024vd3d}
Sherwin Bahmani, Ivan Skorokhodov, Aliaksandr Siarohin, Willi Menapace, Guocheng Qian, Michael Vasilkovsky, Hsin-Ying Lee, Chaoyang Wang, Jiaxu Zou, Andrea Tagliasacchi, et~al.
\newblock Vd3d: Taming large video diffusion transformers for 3d camera control.
\newblock \emph{arXiv preprint arXiv:2407.12781}, 2024{\natexlab{c}}.

\bibitem[Baker et~al.(2007)Baker, Roth, Scharstein, Black, Lewis, and Szeliski]{baker2011database}
Simon Baker, Stefan Roth, Daniel Scharstein, Michael~J. Black, J.P. Lewis, and Richard Szeliski.
\newblock A database and evaluation methodology for optical flow.
\newblock In \emph{2007 IEEE 11th International Conference on Computer Vision}, page 1–8. IEEE, 2007.

\bibitem[Barron et~al.(2022)Barron, Mildenhall, Verbin, Srinivasan, and Hedman]{barron2022mip}
Jonathan~T. Barron, Ben Mildenhall, Dor Verbin, Pratul~P. Srinivasan, and Peter Hedman.
\newblock Mip-{NeRF} 360: Unbounded anti-aliased neural radiance fields.
\newblock In \emph{2022 IEEE/CVF Conference on Computer Vision and Pattern Recognition (CVPR)}. IEEE, 2022.

\bibitem[Blattmann et~al.(2023{\natexlab{a}})Blattmann, Dockhorn, Kulal, Mendelevitch, Kilian, Lorenz, Levi, English, Voleti, Letts, et~al.]{blattmann2023stable}
Andreas Blattmann, Tim Dockhorn, Sumith Kulal, Daniel Mendelevitch, Maciej Kilian, Dominik Lorenz, Yam Levi, Zion English, Vikram Voleti, Adam Letts, et~al.
\newblock Stable video diffusion: Scaling latent video diffusion models to large datasets.
\newblock \emph{arXiv preprint arXiv:2311.15127}, 2023{\natexlab{a}}.

\bibitem[Blattmann et~al.(2023{\natexlab{b}})Blattmann, Rombach, Ling, Dockhorn, Kim, Fidler, and Kreis]{blattmann2023align}
Andreas Blattmann, Robin Rombach, Huan Ling, Tim Dockhorn, Seung~Wook Kim, Sanja Fidler, and Karsten Kreis.
\newblock Align your latents: High-resolution video synthesis with latent diffusion models.
\newblock In \emph{2023 IEEE/CVF Conference on Computer Vision and Pattern Recognition (CVPR)}, page 22563–22575. IEEE, 2023{\natexlab{b}}.

\bibitem[Brooks et~al.(2023)Brooks, Holynski, and Efros]{brooks2023instructpix2pix}
Tim Brooks, Aleksander Holynski, and Alexei~A. Efros.
\newblock Instructpix2pix: Learning to follow image editing instructions.
\newblock In \emph{2023 IEEE/CVF Conference on Computer Vision and Pattern Recognition (CVPR)}, page 18392–18402. IEEE, 2023.

\bibitem[Das et~al.(2024)Das, Wewer, Yunus, Ilg, and Lenssen]{das2023neural}
Devikalyan Das, Christopher Wewer, Raza Yunus, Eddy Ilg, and Jan~Eric Lenssen.
\newblock Neural parametric gaussians for monocular non-rigid object reconstruction.
\newblock In \emph{2024 IEEE/CVF Conference on Computer Vision and Pattern Recognition (CVPR)}, page 10715–10725. IEEE, 2024.

\bibitem[Duisterhof et~al.(2023)Duisterhof, Mandi, Yao, Liu, Shou, Song, and Ichnowski]{duisterhof2023md}
Bardienus~P Duisterhof, Zhao Mandi, Yunchao Yao, Jia-Wei Liu, Mike~Zheng Shou, Shuran Song, and Jeffrey Ichnowski.
\newblock {MD}-splatting: Learning metric deformation from 4d gaussians in highly deformable scenes.
\newblock \emph{arXiv preprint arXiv:2312.00583}, 2023.

\bibitem[Gao et~al.(2024)Gao, Xu, Cao, Mildenhall, Ma, Chen, Tang, and Neumann]{gao2024gaussianflow}
Quankai Gao, Qiangeng Xu, Zhe Cao, Ben Mildenhall, Wenchao Ma, Le Chen, Danhang Tang, and Ulrich Neumann.
\newblock Gaussianflow: Splatting gaussian dynamics for 4d content creation.
\newblock \emph{arXiv preprint arXiv:2403.12365}, 2024.

\bibitem[Haque et~al.(2023)Haque, Tancik, Efros, Holynski, and Kanazawa]{haque2023instructnerf}
Ayaan Haque, Matthew Tancik, Alexei~A. Efros, Aleksander Holynski, and Angjoo Kanazawa.
\newblock Instruct-nerf2nerf: Editing 3d scenes with instructions.
\newblock In \emph{2023 IEEE/CVF International Conference on Computer Vision (ICCV)}, page 19683–19693. IEEE, 2023.

\bibitem[He et~al.(2024)He, Xu, Guo, Wetzstein, Dai, Li, and Yang]{he2024cameractrl}
Hao He, Yinghao Xu, Yuwei Guo, Gordon Wetzstein, Bo Dai, Hongsheng Li, and Ceyuan Yang.
\newblock {CameraCtrl}: Enabling camera control for text-to-video generation, 2024.

\bibitem[Ho et~al.(2020)Ho, Jain, and Abbeel]{ho2020denoising}
Jonathan Ho, Ajay Jain, and Pieter Abbeel.
\newblock Denoising diffusion probabilistic models.
\newblock \emph{Advances in neural information processing systems}, 33:\penalty0 6840--6851, 2020.

\bibitem[Ho et~al.(2022)Ho, Chan, Saharia, Whang, Gao, Gritsenko, Kingma, Poole, Norouzi, Fleet, et~al.]{ho2022imagen}
Jonathan Ho, William Chan, Chitwan Saharia, Jay Whang, Ruiqi Gao, Alexey Gritsenko, Diederik~P Kingma, Ben Poole, Mohammad Norouzi, David~J Fleet, et~al.
\newblock Imagen video: High definition video generation with diffusion models.
\newblock \emph{arXiv preprint arXiv:2210.02303}, 2022.

\bibitem[Huang et~al.(2024{\natexlab{a}})Huang, Zeng, Li, Zuo, and Lau]{huang2024dreamphysics}
Tianyu Huang, Yihan Zeng, Hui Li, Wangmeng Zuo, and Rynson~WH Lau.
\newblock {DreamPhysics}: Learning physical properties of dynamic 3d gaussians with video diffusion priors.
\newblock \emph{arXiv preprint arXiv:2406.01476}, 2024{\natexlab{a}}.

\bibitem[Huang et~al.(2024{\natexlab{b}})Huang, Sun, Yang, Lyu, Cao, and Qi]{huang2023sc}
Yi-Hua Huang, Yang-Tian Sun, Ziyi Yang, Xiaoyang Lyu, Yan-Pei Cao, and Xiaojuan Qi.
\newblock {SC}-{GS}: Sparse-controlled gaussian splatting for editable dynamic scenes.
\newblock In \emph{2024 IEEE/CVF Conference on Computer Vision and Pattern Recognition (CVPR)}, page 4220–4230. IEEE, 2024{\natexlab{b}}.

\bibitem[Jiang et~al.(2024)Jiang, Yu, Cao, Wang, Hu, and Gao]{jiang2024animate3d}
Yanqin Jiang, Chaohui Yu, Chenjie Cao, Fan Wang, Weiming Hu, and Jin Gao.
\newblock Animate3d: Animating any 3d model with multi-view video diffusion.
\newblock \emph{arXiv preprint arXiv:2407.11398}, 2024.

\bibitem[Jung et~al.(2023)Jung, Brasch, Song, Perez-Pellitero, Zhou, Li, Navab, and Busam]{jung2023deformable}
HyunJun Jung, Nikolas Brasch, Jifei Song, Eduardo Perez-Pellitero, Yiren Zhou, Zhihao Li, Nassir Navab, and Benjamin Busam.
\newblock Deformable 3d gaussian splatting for animatable human avatars.
\newblock \emph{arXiv preprint arXiv:2312.15059}, 2023.

\bibitem[Kabsch(1976)]{kabsch1976solution}
W. Kabsch.
\newblock A solution for the best rotation to relate two sets of vectors.
\newblock \emph{Acta Crystallographica Section A}, 32\penalty0 (5):\penalty0 922–923, 1976.

\bibitem[Karaev et~al.(2024)Karaev, Rocco, Graham, Neverova, Vedaldi, and Rupprecht]{karaev2023cotracker}
Nikita Karaev, Ignacio Rocco, Benjamin Graham, Natalia Neverova, Andrea Vedaldi, and Christian Rupprecht.
\newblock Cotracker: It is better to track together.
\newblock In \emph{European Conference on Computer Vision}, 2024.

\bibitem[Kerbl et~al.(2023)Kerbl, Kopanas, Leimkuehler, and Drettakis]{kerbl20233d}
Bernhard Kerbl, Georgios Kopanas, Thomas Leimkuehler, and George Drettakis.
\newblock 3d gaussian splatting for real-time radiance field rendering.
\newblock \emph{ACM Transactions on Graphics}, 42\penalty0 (4):\penalty0 1–14, 2023.

\bibitem[Kuang et~al.(2024)Kuang, Cai, He, Xu, Li, Guibas, and Wetzstein]{kuang2024collaborative}
Zhengfei Kuang, Shengqu Cai, Hao He, Yinghao Xu, Hongsheng Li, Leonidas Guibas, and Gordon Wetzstein.
\newblock Collaborative video diffusion: Consistent multi-video generation with camera control.
\newblock \emph{arXiv preprint arXiv:2405.17414}, 2024.

\bibitem[Lei et~al.(2024)Lei, Weng, Harley, Guibas, and Daniilidis]{lei2024mosca}
Jiahui Lei, Yijia Weng, Adam Harley, Leonidas Guibas, and Kostas Daniilidis.
\newblock {MoSca}: Dynamic gaussian fusion from casual videos via 4d motion scaffolds.
\newblock \emph{arXiv preprint arXiv:2405.17421}, 2024.

\bibitem[Lin et~al.(2023)Lin, Dai, Zhu, and Yao]{lin2023gaussian}
Youtian Lin, Zuozhuo Dai, Siyu Zhu, and Yao Yao.
\newblock Gaussian-flow: 4d reconstruction with dynamic 3d gaussian particle.
\newblock \emph{arXiv preprint arXiv:2312.03431}, 2023.

\bibitem[Ling et~al.(2024)Ling, Kim, Torralba, Fidler, and Kreis]{ling2023align}
Huan Ling, Seung~Wook Kim, Antonio Torralba, Sanja Fidler, and Karsten Kreis.
\newblock Align your gaussians: Text-to-4d with dynamic 3d gaussians and composed diffusion models.
\newblock In \emph{2024 IEEE/CVF Conference on Computer Vision and Pattern Recognition (CVPR)}, page 8576–8588. IEEE, 2024.

\bibitem[Liu et~al.(2023)Liu, Wu, Van~Hoorick, Tokmakov, Zakharov, and Vondrick]{liu2023zero}
Ruoshi Liu, Rundi Wu, Basile Van~Hoorick, Pavel Tokmakov, Sergey Zakharov, and Carl Vondrick.
\newblock Zero-1-to-3: Zero-shot one image to 3d object.
\newblock In \emph{2023 IEEE/CVF International Conference on Computer Vision (ICCV)}, page 9264–9275. IEEE, 2023.

\bibitem[Luiten et~al.(2024)Luiten, Kopanas, Leibe, and Ramanan]{luiten2023dynamic}
Jonathon Luiten, Georgios Kopanas, Bastian Leibe, and Deva Ramanan.
\newblock Dynamic 3d gaussians: Tracking by persistent dynamic view synthesis.
\newblock In \emph{2024 International Conference on 3D Vision (3DV)}, page 800–809. IEEE, 2024.

\bibitem[Mildenhall et~al.(2020)Mildenhall, Srinivasan, Tancik, Barron, Ramamoorthi, and Ng]{mildenhall2021nerf}
Ben Mildenhall, Pratul~P. Srinivasan, Matthew Tancik, Jonathan~T. Barron, Ravi Ramamoorthi, and Ren Ng.
\newblock \emph{{NeRF}: Representing Scenes as Neural Radiance Fields for View Synthesis}, page 405–421.
\newblock Springer International Publishing, 2020.

\bibitem[Müller et~al.(2022)Müller, Evans, Schied, and Keller]{muller2022instant}
Thomas Müller, Alex Evans, Christoph Schied, and Alexander Keller.
\newblock Instant neural graphics primitives with a multiresolution hash encoding.
\newblock \emph{ACM Transactions on Graphics}, 41\penalty0 (4):\penalty0 1–15, 2022.

\bibitem[Niemeyer et~al.(2024)Niemeyer, Manhardt, Rakotosaona, Oechsle, Duckworth, Gosula, Tateno, Bates, Kaeser, and Tombari]{niemeyer2024radsplat}
Michael Niemeyer, Fabian Manhardt, Marie-Julie Rakotosaona, Michael Oechsle, Daniel Duckworth, Rama Gosula, Keisuke Tateno, John Bates, Dominik Kaeser, and Federico Tombari.
\newblock {RadSplat}: Radiance field-informed gaussian splatting for robust real-time rendering with 900+ {FPS}.
\newblock \emph{arXiv preprint arXiv:2403.13806}, 2024.

\bibitem[Pang et~al.(2024)Pang, Zhu, Kortylewski, Theobalt, and Habermann]{pang2023ash}
Haokai Pang, Heming Zhu, Adam Kortylewski, Christian Theobalt, and Marc Habermann.
\newblock {ASH}: Animatable gaussian splats for efficient and photoreal human rendering.
\newblock In \emph{Proceedings of the IEEE/CVF Conference on Computer Vision and Pattern Recognition}, 2024.

\bibitem[Piccinelli et~al.(2024)Piccinelli, Yang, Sakaridis, Segu, Li, Gool, and Yu]{piccinelli2024unidepth}
Luigi Piccinelli, Yung-Hsu Yang, Christos Sakaridis, Mattia Segu, Siyuan Li, Luc~Van Gool, and Fisher Yu.
\newblock {UniDepth}: Universal monocular metric depth estimation.
\newblock In \emph{2024 IEEE/CVF Conference on Computer Vision and Pattern Recognition (CVPR)}, page 10106–10116. IEEE, 2024.

\bibitem[Poole et~al.(2022)Poole, Jain, Barron, and Mildenhall]{poole2022dreamfusion}
Ben Poole, Ajay Jain, Jonathan~T Barron, and Ben Mildenhall.
\newblock {DreamFusion}: Text-to-3d using 2d diffusion.
\newblock In \emph{The Eleventh International Conference on Learning Representations}, 2022.

\bibitem[Qin et~al.(2024)Qin, Li, Zhou, Wang, and Pfister]{qin2023langsplat}
Minghan Qin, Wanhua Li, Jiawei Zhou, Haoqian Wang, and Hanspeter Pfister.
\newblock {LangSplat}: 3d language gaussian splatting.
\newblock In \emph{2024 IEEE/CVF Conference on Computer Vision and Pattern Recognition (CVPR)}, page 20051–20060. IEEE, 2024.

\bibitem[Radford et~al.(2021)Radford, Kim, Hallacy, Ramesh, Goh, Agarwal, Sastry, Askell, Mishkin, Clark, et~al.]{radford2021learning}
Alec Radford, Jong~Wook Kim, Chris Hallacy, Aditya Ramesh, Gabriel Goh, Sandhini Agarwal, Girish Sastry, Amanda Askell, Pamela Mishkin, Jack Clark, et~al.
\newblock Learning transferable visual models from natural language supervision.
\newblock In \emph{International conference on machine learning}, pages 8748--8763. PMLR, 2021.

\bibitem[Ren et~al.(2023)Ren, Pan, Tang, Zhang, Cao, Zeng, and Liu]{ren2023dreamgaussian4d}
Jiawei Ren, Liang Pan, Jiaxiang Tang, Chi Zhang, Ang Cao, Gang Zeng, and Ziwei Liu.
\newblock Dreamgaussian4d: Generative 4d gaussian splatting.
\newblock \emph{arXiv preprint arXiv:2312.17142}, 2023.

\bibitem[Ren et~al.(2024)Ren, Xie, Mirzaei, Liang, Zeng, Kreis, Liu, Torralba, Fidler, Kim, et~al.]{ren2024l4gm}
Jiawei Ren, Kevin Xie, Ashkan Mirzaei, Hanxue Liang, Xiaohui Zeng, Karsten Kreis, Ziwei Liu, Antonio Torralba, Sanja Fidler, Seung~Wook Kim, et~al.
\newblock L4gm: Large 4d gaussian reconstruction model.
\newblock \emph{arXiv preprint arXiv:2406.10324}, 2024.

\bibitem[Rombach et~al.(2022)Rombach, Blattmann, Lorenz, Esser, and Ommer]{rombach2022high}
Robin Rombach, Andreas Blattmann, Dominik Lorenz, Patrick Esser, and Bjorn Ommer.
\newblock High-resolution image synthesis with latent diffusion models.
\newblock In \emph{2022 IEEE/CVF Conference on Computer Vision and Pattern Recognition (CVPR)}, page 10674–10685. IEEE, 2022.

\bibitem[Shao et~al.(2024)Shao, Wang, Li, Wang, Lin, Zhang, Fan, and Wang]{shao2024splattingavatar}
Zhijing Shao, Zhaolong Wang, Zhuang Li, Duotun Wang, Xiangru Lin, Yu Zhang, Mingming Fan, and Zeyu Wang.
\newblock {SplattingAvatar}: Realistic real-time human avatars with mesh-embedded gaussian splatting.
\newblock In \emph{2024 IEEE/CVF Conference on Computer Vision and Pattern Recognition (CVPR)}, page 1606–1616. IEEE, 2024.

\bibitem[Shi et~al.(2023)Shi, Wang, Ye, Mai, Li, and Yang]{shi2023mvdream}
Yichun Shi, Peng Wang, Jianglong Ye, Long Mai, Kejie Li, and Xiao Yang.
\newblock {MVDream}: Multi-view diffusion for 3d generation.
\newblock In \emph{The Twelfth International Conference on Learning Representations}, 2023.

\bibitem[Singer et~al.(2022)Singer, Polyak, Hayes, Yin, An, Zhang, Hu, Yang, Ashual, Gafni, et~al.]{singer2022make}
Uriel Singer, Adam Polyak, Thomas Hayes, Xi Yin, Jie An, Songyang Zhang, Qiyuan Hu, Harry Yang, Oron Ashual, Oran Gafni, et~al.
\newblock Make-a-video: Text-to-video generation without text-video data.
\newblock \emph{arXiv preprint arXiv:2209.14792}, 2022.

\bibitem[Singer et~al.(2023)Singer, Sheynin, Polyak, Ashual, Makarov, Kokkinos, Goyal, Vedaldi, Parikh, Johnson, et~al.]{singer2023text}
Uriel Singer, Shelly Sheynin, Adam Polyak, Oron Ashual, Iurii Makarov, Filippos Kokkinos, Naman Goyal, Andrea Vedaldi, Devi Parikh, Justin Johnson, et~al.
\newblock Text-to-4d dynamic scene generation.
\newblock In \emph{Proceedings of the 40th International Conference on Machine Learning}, pages 31915--31929, 2023.

\bibitem[Stearns et~al.(2024)Stearns, Harley, Uy, Dubost, Tombari, Wetzstein, and Guibas]{stearns2024dynamic}
Colton Stearns, Adam Harley, Mikaela Uy, Florian Dubost, Federico Tombari, Gordon Wetzstein, and Leonidas Guibas.
\newblock Dynamic gaussian marbles for novel view synthesis of casual monocular videos.
\newblock \emph{arXiv preprint arXiv:2406.18717}, 2024.

\bibitem[Tancik et~al.(2020)Tancik, Srinivasan, Mildenhall, Fridovich-Keil, Raghavan, Singhal, Ramamoorthi, Barron, and Ng]{tancik2020fourier}
Matthew Tancik, Pratul Srinivasan, Ben Mildenhall, Sara Fridovich-Keil, Nithin Raghavan, Utkarsh Singhal, Ravi Ramamoorthi, Jonathan Barron, and Ren Ng.
\newblock Fourier features let networks learn high frequency functions in low dimensional domains.
\newblock \emph{Advances in Neural Information Processing Systems}, 33:\penalty0 7537--7547, 2020.

\bibitem[Tang et~al.(2024)Tang, Chen, Chen, Wang, Zeng, and Liu]{tang2024lgm}
Jiaxiang Tang, Zhaoxi Chen, Xiaokang Chen, Tengfei Wang, Gang Zeng, and Ziwei Liu.
\newblock Lgm: Large multi-view gaussian model for high-resolution 3d content creation.
\newblock \emph{arXiv preprint arXiv:2402.05054}, 2024.

\bibitem[Teed and Deng(2021)]{teed2020raft}
Zachary Teed and Jia Deng.
\newblock {RAFT}: Recurrent all-pairs field transforms for optical flow (extended abstract).
\newblock In \emph{Proceedings of the Thirtieth International Joint Conference on Artificial Intelligence}, page 4839–4843. International Joint Conferences on Artificial Intelligence Organization, 2021.

\bibitem[Umeyama(1991)]{umeyama1991least}
S. Umeyama.
\newblock Least-squares estimation of transformation parameters between two point patterns.
\newblock \emph{IEEE Transactions on Pattern Analysis and Machine Intelligence}, 13\penalty0 (4):\penalty0 376–380, 1991.

\bibitem[Wang et~al.(2023{\natexlab{a}})Wang, Du, Li, Yeh, and Shakhnarovich]{wang2023score}
Haochen Wang, Xiaodan Du, Jiahao Li, Raymond~A. Yeh, and Greg Shakhnarovich.
\newblock Score jacobian chaining: Lifting pretrained 2d diffusion models for 3d generation.
\newblock In \emph{2023 IEEE/CVF Conference on Computer Vision and Pattern Recognition (CVPR)}, page 12619–12629. IEEE, 2023{\natexlab{a}}.

\bibitem[Wang et~al.(2023{\natexlab{b}})Wang, Yuan, Chen, Zhang, Wang, and Zhang]{wang2023modelscope}
Jiuniu Wang, Hangjie Yuan, Dayou Chen, Yingya Zhang, Xiang Wang, and Shiwei Zhang.
\newblock Modelscope text-to-video technical report.
\newblock \emph{arXiv preprint arXiv:2308.06571}, 2023{\natexlab{b}}.

\bibitem[Wang et~al.(2023{\natexlab{c}})Wang, Yuan, Zhang, Chen, Wang, Zhang, Shen, Zhao, and Zhou]{wang2023videocomposer}
Xiang Wang, Hangjie Yuan, Shiwei Zhang, Dayou Chen, Jiuniu Wang, Yingya Zhang, Yujun Shen, Deli Zhao, and Jingren Zhou.
\newblock {VideoComposer}: Compositional video synthesis with motion controllability.
\newblock \emph{arXiv e-prints}, pages arXiv--2306, 2023{\natexlab{c}}.

\bibitem[Wang et~al.(2024)Wang, Lu, Wang, Bao, Li, Su, and Zhu]{wang2024prolificdreamer}
Zhengyi Wang, Cheng Lu, Yikai Wang, Fan Bao, Chongxuan Li, Hang Su, and Jun Zhu.
\newblock Prolificdreamer: High-fidelity and diverse text-to-3d generation with variational score distillation.
\newblock \emph{Advances in Neural Information Processing Systems}, 36, 2024.

\bibitem[Wu et~al.(2024{\natexlab{a}})Wu, Yi, Fang, Xie, Zhang, Wei, Liu, Tian, and Wang]{wu20234d}
Guanjun Wu, Taoran Yi, Jiemin Fang, Lingxi Xie, Xiaopeng Zhang, Wei Wei, Wenyu Liu, Qi Tian, and Xinggang Wang.
\newblock 4d gaussian splatting for real-time dynamic scene rendering.
\newblock In \emph{2024 IEEE/CVF Conference on Computer Vision and Pattern Recognition (CVPR)}, page 20310–20320. IEEE, 2024{\natexlab{a}}.

\bibitem[Wu et~al.(2024{\natexlab{b}})Wu, Mildenhall, Henzler, Park, Gao, Watson, Srinivasan, Verbin, Barron, Poole, and Hołyński]{wu2023reconfusion}
Rundi Wu, Ben Mildenhall, Philipp Henzler, Keunhong Park, Ruiqi Gao, Daniel Watson, Pratul~P. Srinivasan, Dor Verbin, Jonathan~T. Barron, Ben Poole, and Aleksander Hołyński.
\newblock {ReconFusion}: 3d reconstruction with diffusion priors.
\newblock In \emph{2024 IEEE/CVF Conference on Computer Vision and Pattern Recognition (CVPR)}, page 21551–21561. IEEE, 2024{\natexlab{b}}.

\bibitem[Xie et~al.(2024{\natexlab{a}})Xie, Zong, Qiu, Li, Feng, Yang, and Jiang]{xie2023physgaussian}
Tianyi Xie, Zeshun Zong, Yuxing Qiu, Xuan Li, Yutao Feng, Yin Yang, and Chenfanfu Jiang.
\newblock {PhysGaussian}: Physics-integrated 3d gaussians for generative dynamics.
\newblock In \emph{2024 IEEE/CVF Conference on Computer Vision and Pattern Recognition (CVPR)}, page 4389–4398. IEEE, 2024{\natexlab{a}}.

\bibitem[Xie et~al.(2024{\natexlab{b}})Xie, Yao, Voleti, Jiang, and Jampani]{xie2024sv4d}
Yiming Xie, Chun-Han Yao, Vikram Voleti, Huaizu Jiang, and Varun Jampani.
\newblock Sv4d: Dynamic 3d content generation with multi-frame and multi-view consistency.
\newblock \emph{arXiv preprint arXiv:2407.17470}, 2024{\natexlab{b}}.

\bibitem[Xing et~al.(2024)Xing, Xia, Zhang, Chen, Yu, Liu, Wang, Wong, and Shan]{xing2023dynamicrafter}
Jinbo Xing, Menghan Xia, Yong Zhang, Haoxin Chen, Wangbo Yu, Hanyuan Liu, Xintao Wang, Tien-Tsin Wong, and Ying Shan.
\newblock {DynamiCrafter}: Animating open-domain images with video diffusion priors.
\newblock In \emph{European Conference on Computer Vision}, 2024.

\bibitem[Yang et~al.(2024)Yang, Gao, Zhou, Jiao, Zhang, and Jin]{yang2023deformable}
Ziyi Yang, Xinyu Gao, Wen Zhou, Shaohui Jiao, Yuqing Zhang, and Xiaogang Jin.
\newblock Deformable 3d gaussians for high-fidelity monocular dynamic scene reconstruction.
\newblock In \emph{2024 IEEE/CVF Conference on Computer Vision and Pattern Recognition (CVPR)}, page 20331–20341. IEEE, 2024.

\bibitem[Yu et~al.(2024{\natexlab{a}})Yu, Wang, Zhuang, Menapace, Siarohin, Cao, Jeni, Tulyakov, and Lee]{yu20244real}
Heng Yu, Chaoyang Wang, Peiye Zhuang, Willi Menapace, Aliaksandr Siarohin, Junli Cao, Laszlo~A Jeni, Sergey Tulyakov, and Hsin-Ying Lee.
\newblock 4real: Towards photorealistic 4d scene generation via video diffusion models.
\newblock \emph{arXiv preprint arXiv:2406.07472}, 2024{\natexlab{a}}.

\bibitem[Yu et~al.(2024{\natexlab{b}})Yu, Guo, Li, Liang, Zhang, and Qi]{yu2023text}
Xin Yu, Yuan-Chen Guo, Yangguang Li, Ding Liang, Song-Hai Zhang, and Xiaojuan Qi.
\newblock Text-to-3d with classifier score distillation.
\newblock In \emph{The Thirteenth International Conference on Learning Representations}, 2024{\natexlab{b}}.

\bibitem[Zhang et~al.(2024)Zhang, Yu, Wu, Feng, Zheng, Snavely, Wu, and Freeman]{zhang2024physdreamer}
Tianyuan Zhang, Hong-Xing Yu, Rundi Wu, Brandon~Y Feng, Changxi Zheng, Noah Snavely, Jiajun Wu, and William~T Freeman.
\newblock {PhysDreamer}: Physics-based interaction with 3d objects via video generation.
\newblock \emph{arXiv preprint arXiv:2404.13026}, 2024.

\bibitem[Zhou and Tulsiani(2023)]{zhou2023sparsefusion}
Zhizhuo Zhou and Shubham Tulsiani.
\newblock {SparseFusion}: Distilling view-conditioned diffusion for 3d reconstruction.
\newblock In \emph{2023 IEEE/CVF Conference on Computer Vision and Pattern Recognition (CVPR)}, page 12588–12597. IEEE, 2023.

\end{thebibliography}
}

\makeatletter
\setcounter{section}{-1}\stepcounter{section}
\setcounter{figure}{-1}\stepcounter{figure}
\renewcommand \thesection{S\@arabic\c@section}
\renewcommand\thetable{S\@arabic\c@table}
\renewcommand \thefigure{S\@arabic\c@figure}
\makeatother

\clearpage
\setcounter{page}{1}
\maketitlesupplementary

\mysection{Implementation Details}
We make our code available under the following URL: \url{https://github.com/wimmerth/gaussians2life}. Optimization of dynamics in a single scene takes about 10 minutes on average. In our experiments, video generation is limited to 8 frames to balance computational constraints with the requirements of the video diffusion model. We note that this also limits the amount of motion that can be effectively distilled into the scene. 

\begin{figure}[ht]
    \centering
    \includegraphics[width=\columnwidth]{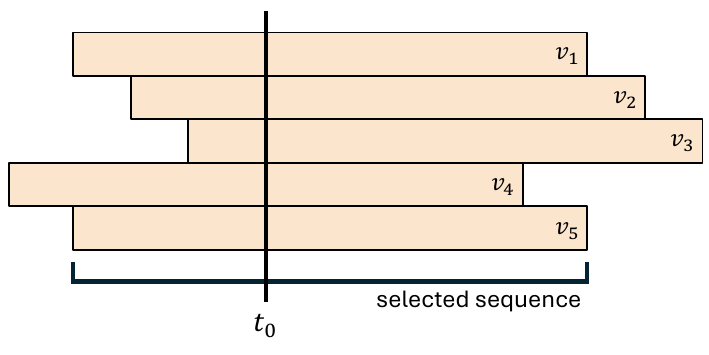}
    \caption{Temporal alignment of predicted trajectories from different viewpoints at $t_0$. The sequence used as input to the next optimization step is the sequence with the most overlap between the source trajectories.}
    \label{fig:temporal-alignment}
    \vspace{-.5\intextsep}
\end{figure}

\mysubsection{Viewpoint Sampling} \label{app:viewpoint-sampling}
Our proposed approach for approximately multi-view consistent video generation is based on the assumption that there are no large viewpoint changes between subsequent optimization steps. The viewpoint sampling strategy has to reflect this constraint while still enabling guidance from views all around the object. To do so, we start at an anchor viewpoint and sample viewpoints on the sphere around the scene center, where we first sample two endpoints lying at the maximum azimuth change $c_{\textrm{azm,anchor}} \pm m_{\textrm{azm}}$, with an elevation deviating from $c_{\textrm{elv,anchor}}$ by at most $m_{\textrm{elv}}$, and a distance change of at most $m_{\textrm{dist}} \cdot c_{\textrm{dist,anchor}}$, both sampled uniformly. We then spread out $n_{\textrm{c}}$ views uniformly along the path on the sphere from the anchor point to each endpoint. Finally, we disturb each point's azimuth, elevation, and distance with a small noise sampled using standard deviations $\sigma_{\textrm{azm}}, \sigma_{\textrm{elv}}, \sigma_{\textrm{dist}}$.

\mysubsection{Warping using Optical Flow}\label{app:flow-warping}

Before feeding the previously generated video $v_{s-1}$ through the video diffusion model encoder and using it as motion prior for the generation of a video from the new viewpoint (see \cref{fig:mv-diffusion}), we aim to warp it to resemble the looks of a video taken from the new viewpoint as closely as possible.

To do so, one possible approach is to estimate correspondences between the renderings of the given, static 3D scene. Using these correspondences, the optimal homography can be computed, which can subsequently used for warping all frames of the previous video using a perspective projection. While this approach seems reasonable, it comes with the limitation that larger camera pose changes will result in heavy distortions and unrealistic results. Additionally, fore- and background objects are not taken care of separately.

\begin{figure}
    \centering
    \includegraphics[width=\columnwidth]{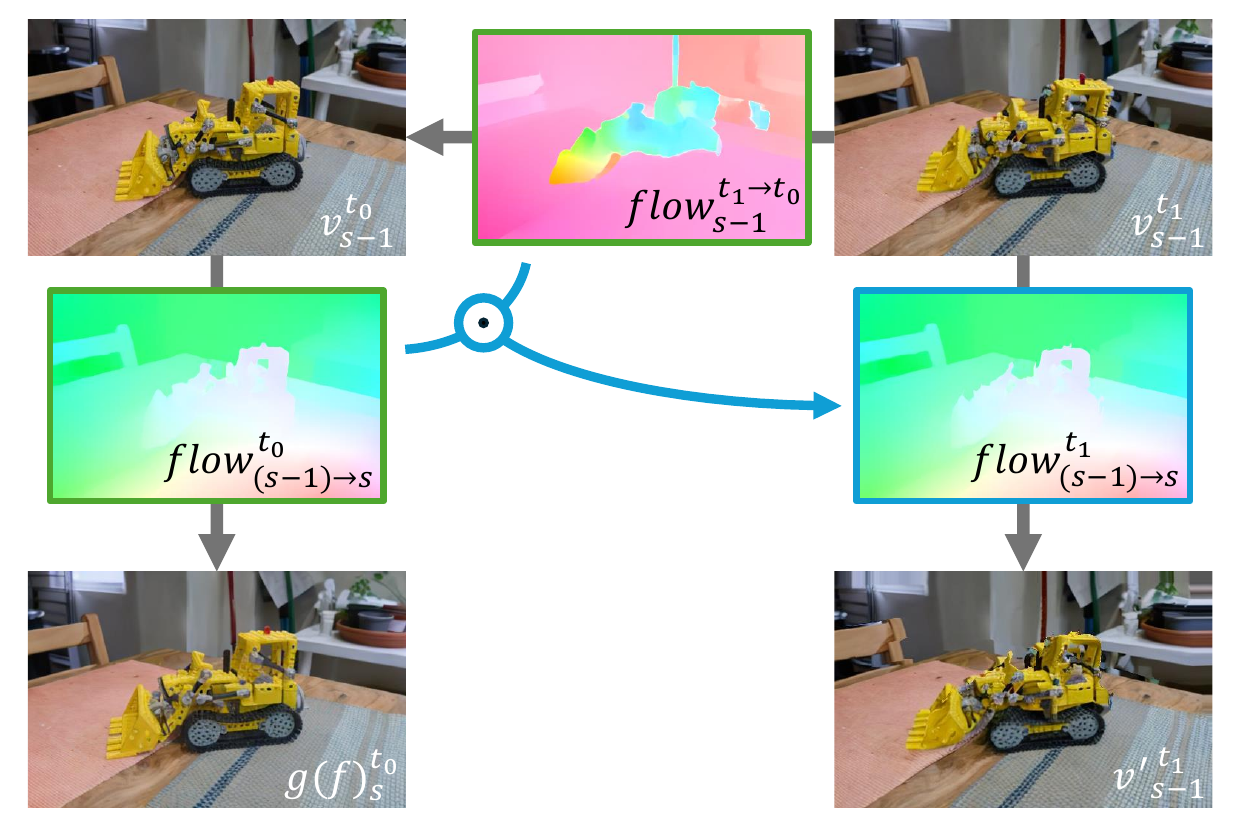}
    \caption{Warping of previously generated video frames $v_{s-1}$ to resemble a video captured from the viewpoint at the next optimization step $s$. First, \textcolor{flowgreen}{optical flow} between the video frames of $v_{s-1}$ is computed that is subsequently composed with the optical flow computed between the renderings of the static 3D scene from the two viewpoints at $s-1$ and $s$. The \textcolor{flowblue}{composed flow} is then used to warp the video frames of $v_{s-1}^{t_i}$ to their respective equivalents in the new video ${v^{\prime}}_{s-1}^{t_i}$ that is used as motion prior for the video diffusion model.}
    \label{fig:flow-warping}
\end{figure}

To resolve these issues, we make use of an off-the-shelf optical flow estimation method~\citep{teed2020raft}, which we use to compute the optical flow $flow_{(s-1)\rightarrow s}^{t_0}$ between the static scene renderings at steps $s-1$ and $s$, as well as the optical flow $flow_{s-1}^{t_i \rightarrow t_0}$ from other frames in the previous video $v_{s-1}^{t_{i\neq 0}}$ to the static scene rendering $v_{s-1}^{t_0}$.
Finally, we remap the optical flow $flow_{(s-1)\rightarrow s}^{t_0}$ using $flow_{s-1}^{t_i \rightarrow t_0}$ to get $flow_{(s-1) \rightarrow s}^{t_i}$, which is used to warp video frame $v_{s-1}^{t_i}$, see \cref{fig:flow-warping}.  

\mysubsection{Optimization-based Baseline} \label{sec:supp-baseline-ours}
We model the deformations of the scene by optimizing a neural network $f_\theta$ that maps input coordinates $x \in \R^3$ and a time $t \in \R$ to a change in position $\delta x \in \R^3$, and optionally a change in rotation $\delta q \in \R^4$ and scaling $\delta s \in \R^3$. We thus use the 3D scene initialization as canonical representation, i.e.,~the displacement, rotation, or scaling of a Gaussian at position $x = \mu_i$ in the static 3D scene at time $t$ is obtained through querying the deformation field $f_\theta(x,t)$.

We employ a multi-resolution hash encoding for spatial input coordinates $x$~\citep{muller2022instant}, as well as random Fourier features as encoding for $t$~\citep{tancik2020fourier}. We use a standard MLP with separate heads for the prediction of displacement, rotation, and scale changes, where we multiply outputs with $t^{0.35}$, following previous works~\citep{ling2023align}, to fix the scene initialization at $t=0$.

\begin{figure}
    \centering
    \includegraphics[width=\linewidth]{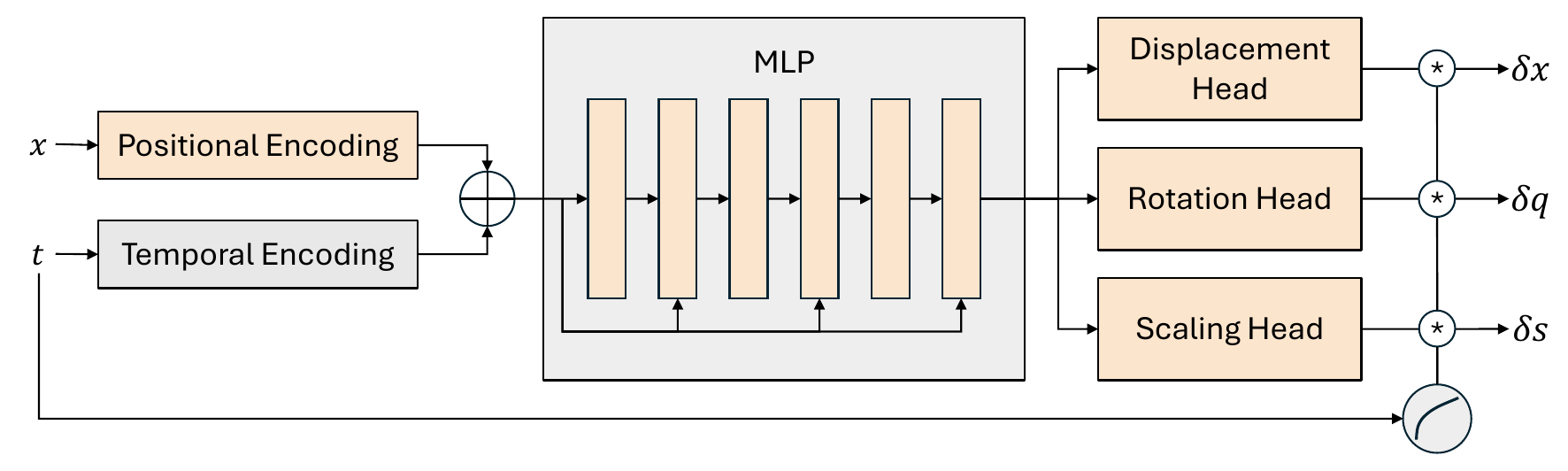}
    \caption{Architecture of the neural deformation field used in our optimization-based baseline. Learnable components in \textcolor{boxorange}{orange}.}
    \label{fig:neural-field-baseline}
    \vspace{-.5\intextsep}
\end{figure}

We further use a similar deformation technique as presented in \cref{sec:motion-transfer}, where we first sample a set of anchor points using $k$-Means, that are used to query the deformation field, and motion is subsequently transferred to all 3D Gaussians using the presented techniques in \cref{sec:motion-transfer} to get smoother deformations.

We use our approximately multi-view consistent videos as guidance, where we employ standard rendering-based losses, i.e.~$L_1$, D-SSIM, and LPIPS, as well as several regularization terms to promote rigid motion~\citep{huang2023sc} and preserve momentum~\citep{duisterhof2023md}, as well as local isometries~\citep{luiten2023dynamic}. Additionally, we make use of an optical flow supervision signal following \citet{gao2024gaussianflow}.

\mysubsection{DreamGaussian4D Baseline} \label{sec:supp-baseline-dreamgaussian4d}
DreamGaussian4D~\citep{ren2023dreamgaussian4d} is a video-to-4D method that works by first generating a static 3D asset using multi-view SDS or the feed-forward LGM model~\citep{tang2024lgm}. This 3D model is then deformed, guided by SDS with a multi-view image diffusion model that is conditioned on the frames of a guidance video from an anchor viewpoint. To enable a comparison to our method, we use the same 3DGS initialization and the same first sampled diffusion videos for both methods. Further, if a mask for the moving 3D objects is given, we apply it to the 3DGS scene and only use this object as input to DreamGaussian4D, which is necessary as the method is based only on single objects without backgrounds and will otherwise collapse the backgrounds. At test time, we can then add the background Gaussians back into the 3D scene.
Importantly, we do not use the second stage of their pipeline, which extracts per-frame meshes and further refines textures on them using SDS with a video diffusion model, as we cannot integrate 3D meshes into the full 3DGS scene.

\mysubsection{Hyperparameters} \label{sec:supp-baseline-hyperparameters}
We linearly decrease the noise level from 0.75 to 0.2 for the diffusion model inputs and similarly decrease the latent interpolation weight $\lambda_{\textrm{previous}}$ from 0.6 to 0.0. We use 40 de-noising steps for video generation from new viewpoints. We estimate the motion of every 3D Gaussian from its nearest $n \in [50, 150]$ anchor trajectories, where we increase $n$ over time as more and more anchor trajectories are added. For this, we generally use a high temperature $\tau$ (\cref{eq:method-linear-interpolation}) but experiment with different weighting strategies. We note that we tracked 1600 points per video originally, but this number is usually reduced to about 600 valid trajectories that lie within the object bounding box at $t_0$ and that have consistent tracking. Due to the high memory consumption of the used video diffusion model and limited computing resources, we limit our experiments to the generation of $n=8$ frame guidance videos.

In the comparison against our baseline method, we optimize the deformation field for 14,000 steps. We increase the regularization level over the duration of the optimization and use both the rendering- and the optical flow-based losses. In addition, we reduce the latent interpolation from 0.5 to 0.0 and the noise level from 0.75 to 0.4 throughout the optimization. Every 4,000 steps, we sample new guidance videos from 15 viewpoints and alternate between these viewpoints during the optimization.

\mysection{Additional Results}

\begin{figure}
    \centering
    \includegraphics[width=.4\columnwidth]{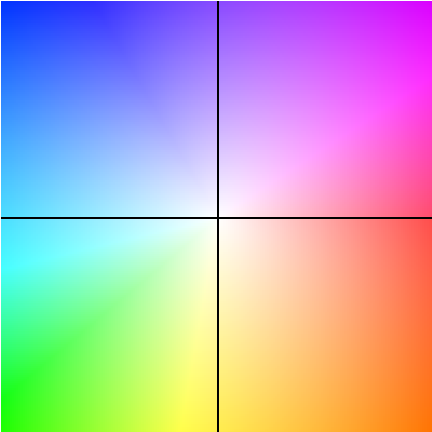}
    \caption{Colormap used for flow visualization in \cref{fig:qualitative-results} as proposed by \citet{baker2011database}.}
    \label{fig:flow-cmap}
\end{figure}

\mysubsection{Quantitative Ablation} \label{sec:quantitative-ablation}

\begin{table*}[htb!]
\caption{Quantitative results for several ablated versions of our method on the Mip-NeRF 360 LEGO bulldozer scene. We note that this evaluation mainly shows the insufficiency of any single quantitative metric. The qualitatively best result (see \cref{fig:qualitative-results}) does not outperform the other versions on any single one of the metrics but does perform the most consistently across all categories.}
\label{tab:quantitative-results-lego}
\centering
\begin{tabular}{l|rrrrrr}
Metric & \multicolumn{1}{c}{Ours} & \multicolumn{1}{c}{\begin{tabular}[c]{@{}c@{}}No MV\\ Diffusion\end{tabular}} & \multicolumn{1}{c}{\begin{tabular}[c]{@{}c@{}}No 2D-3D\\ lifting\end{tabular}} & \multicolumn{1}{c}{\begin{tabular}[c]{@{}c@{}}Rigid motion\\ estimation\end{tabular}} & \multicolumn{1}{c}{\begin{tabular}[c]{@{}c@{}}Fewer inter-\\polation anchors\end{tabular}} & \multicolumn{1}{c}{\begin{tabular}[c]{@{}c@{}}More inter-\\polation anchors\end{tabular}} \\ \hline
\scriptsize{Motion Amount (rank) } & \scriptsize{ 3 } & \scriptsize{ 4 } & \scriptsize{ 5 } & \scriptsize{ 2 } & \scriptsize{ 1 } & \scriptsize{ 6 }\\
Displacement ($10^{-4}$) & \cellcolor[HTML]{FED666}{\color[HTML]{434343} 1.84} & \cellcolor[HTML]{FED567}{\color[HTML]{434343} 1.83} & \cellcolor[HTML]{FDD067}{\color[HTML]{434343} 1.82} & \cellcolor[HTML]{86C37F}{\color[HTML]{434343} 2.02} & \cellcolor[HTML]{57BB8A}{\color[HTML]{434343} \textbf{2.09}} & \cellcolor[HTML]{E67C73}{\color[HTML]{434343} 1.54} \\ \hline
\scriptsize{Geometry / Physics ($\varnothing$ rank) } & \scriptsize{ 3.00 } & \scriptsize{ 2.50 } & \scriptsize{ 4.75 } & \scriptsize{ 4.25 } & \scriptsize{ 5.25 } & \scriptsize{ 1.25 }\\
Rigidity ($10^{-5}$) & \cellcolor[HTML]{BACA75}{\color[HTML]{434343} 1.70} & \cellcolor[HTML]{99C57C}{\color[HTML]{434343} 1.52} & \cellcolor[HTML]{E67C73}{\color[HTML]{434343} 5.12} & \cellcolor[HTML]{FCCB67}{\color[HTML]{434343} 2.46} & \cellcolor[HTML]{F6B46A}{\color[HTML]{434343} 3.24} & \cellcolor[HTML]{57BB8A}{\color[HTML]{434343} \textbf{1.15}} \\
Momentum ($10^{-5}$) & \cellcolor[HTML]{CCCD71}{\color[HTML]{434343} 3.92} & \cellcolor[HTML]{DFD06D}{\color[HTML]{434343} 4.02} & \cellcolor[HTML]{E67C73}{\color[HTML]{434343} 14.44} & \cellcolor[HTML]{FFD566}{\color[HTML]{434343} 4.32} & \cellcolor[HTML]{FED366}{\color[HTML]{434343} 4.62} & \cellcolor[HTML]{57BB8A}{\color[HTML]{434343} \textbf{3.36}} \\
Isometry ($10^{-5}$) & \cellcolor[HTML]{AEC978}{\color[HTML]{434343} 3.11} & \cellcolor[HTML]{99C57C}{\color[HTML]{434343} 2.85} & \cellcolor[HTML]{E67C73}{\color[HTML]{434343} 6.92} & \cellcolor[HTML]{F7B76A}{\color[HTML]{434343} 5.10} & \cellcolor[HTML]{F1A36D}{\color[HTML]{434343} 5.72} & \cellcolor[HTML]{57BB8A}{\color[HTML]{434343} \textbf{2.03}} \\
Rotation similarity ($10^{-4}$) & \cellcolor[HTML]{FED066}{\color[HTML]{434343} 4.00} & \cellcolor[HTML]{F0D36A}{\color[HTML]{434343} 3.35} & \cellcolor[HTML]{57BB8A}{\color[HTML]{434343} \textbf{0.01}} & \cellcolor[HTML]{FCC967}{\color[HTML]{434343} 4.31} & \cellcolor[HTML]{E67C73}{\color[HTML]{434343} 8.04} & \cellcolor[HTML]{D0CE71}{\color[HTML]{434343} 2.65} \\ \hline
\scriptsize{Appearance ($\varnothing$ rank) } & \scriptsize{ 2.50 } & \scriptsize{ 3.50 } & \scriptsize{ 6.00 } & \scriptsize{ 3.00 } & \scriptsize{ 3.00 } & \scriptsize{ 3.00 } \\
$\textrm{CLIP}_{\textrm{text}}$ ($10^{-2}$)& \cellcolor[HTML]{DBD16D}{\color[HTML]{434343} 33.06} & \cellcolor[HTML]{FED567}{\color[HTML]{434343} 33.02} & \cellcolor[HTML]{E67C73}{\color[HTML]{434343} 29.48} & \cellcolor[HTML]{57BB8A}{\color[HTML]{434343} \textbf{33.14}} & \cellcolor[HTML]{CDCE70}{\color[HTML]{434343} 33.07} & \cellcolor[HTML]{FCCC68}{\color[HTML]{434343} 32.65} \\
$\textrm{CLIP}_{\textrm{temporal}}$ ($10^{-2}$)& \cellcolor[HTML]{C6CD72}{\color[HTML]{434343} 99.68} & \cellcolor[HTML]{DFD16C}{\color[HTML]{434343} 99.67} & \cellcolor[HTML]{E67C73}{\color[HTML]{434343} 96.90} & \cellcolor[HTML]{FED567}{\color[HTML]{434343} 99.63} & \cellcolor[HTML]{FED567}{\color[HTML]{434343} 99.64} & \cellcolor[HTML]{57BB8A}{\color[HTML]{434343} \textbf{99.72}} \\ \hline
\scriptsize{Rank over all categories} & \scriptsize{ \textbf{1 (2.83)} } & \scriptsize{ 4 (3.33) } & \scriptsize{ 6 (5.25) } & \scriptsize{ 2 (3.08) } & \scriptsize{ 2 (3.08) } & \scriptsize{ 5 (3.42)}
\end{tabular}
\end{table*}

The quantitative evaluation of generative models is challenging due to missing ground truth. While comparisons with target data distributions can sometimes be computed for 2D or 3D generative models, this is impossible for our method, which would require a collection of ground-truth motions for a given 3D scene. In our case, we believe that a possible metric that can be used is the CLIP similarity score, which we present and extend for measuring coherence among video frames in the next paragraph.

\myparagraph{CLIP Similarity}
\citet{radford2021learning} presented CLIP, a vision-language model that is trained to align the representations of a vision encoder $\textrm{CLIP}_V$ with those of a text encoder $\textrm{CLIP}_T$. It is possible to use this model as an evaluation method, as we can take the cosine similarity computed between the features from the image and the text encoder for generated 2D outputs given a text prompt.

A straightforward metric to measure text alignment is the averaging over multiple frames taken from multiple viewpoints:
\begin{equation} \label{eq:clip-score-simple}
\begin{aligned}
    &\textrm{CLIP}_{\textrm{text}}(f, s, p) = \E_{v \sim \mathcal{V}, t \sim [0,1]}\\
    &\quad\left[\cos\left(\angle(\textrm{CLIP}_V(g(f(s,p);v,t)), \textrm{CLIP}_T(p))\right)\right],
\end{aligned}
\end{equation}
where $f$ is the optimized 4D-scene taking a static 3D scene $s$ and a text prompt $p$ as input, and $g(\cdot;v,t)$ is the rendering of the 4D-scene at timestep $t$ and from viewpoint $v$. 

In the following, we propose a second metric to capture the temporal coherence of rendered videos. For that, we average the similarity of the extracted image embeddings over pairs of successive video frames. Coherency in successive frames can thus be measured, which also is a good measure for the temporal coherence of the full video when averaged over all frame pairs of the generated video:
\begin{equation} \label{eq:clip-score-temporal}
\begin{aligned}
&\textrm{CLIP}_{\textrm{temporal}}(f, s, p) = \E_{v \sim \mathcal{V}, t \sim [0,1)}\\
&\quad[\cos(\angle(\textrm{CLIP}_V(g(f(s,p);v,t)), \\
&\quad\quad\textrm{CLIP}_V(g(f(s,p);v,t+\delta t))))],
\end{aligned}
\end{equation}
where $\delta t$ is the difference between two video frames in practice. We note that this metric is maximized with static scene renderings, which need to be considered during evaluation.

\myparagraph{Regularization Terms as Metrics}
Besides evaluating the visual quality and appearance of dynamic scenes using CLIP losses, we argue that reporting the scores for some of the regularization terms can be used to determine the actual temporal and geometric consistency of scenes. As such, regularization terms are based on the idea of steering the optimization towards animations that are more plausible; they can also be used to validate the optimized results. 

In our evaluation in \cref{tab:quantitative-results-lego}, we thus report the scores for four regularization terms proposed by previous works, that measure local rigidity~\citep{luiten2023dynamic}, momentum preservation~\citep{duisterhof2023md}, local isometry preservation~\citep{luiten2023dynamic}, and local rotation similarity~\citep{luiten2023dynamic}. We note, however, that these scores are minimized (best possible result) for zero motion. Thus, these metrics are also not sufficient as stand-alone tools to quantify the quality of generated motion.

Finally, we also report the average displacement of the 3D Gaussians between all timesteps. As we realize that bringing motion into the scenes is often challenging, we mark higher values for these categories as better. However, as before, this metric alone can not quantify the quality of the generated motion, as a diverging scene, where 3D Gaussians are moved freely in the space, would have a very high value.

As can be seen in \cref{tab:quantitative-results-lego}, the qualitatively best result does not top the quantitative rankings for any single metric. It is, however, the best method when averaging performance across all three evaluation categories: motion amount, geometry and physics, and appearance.

Analyzing the results further, we can see parallels with our qualitative analysis. When no multi-view consistent diffusion is used, performance deteriorates slightly, while using an optimization-based baseline leads to significantly worse results. Contrarily, the results when using rigid motion estimation are worse in terms of geometric metrics, i.e.,~rigidity, than when using linear interpolation, as analyzed in \cref{sec:qualitative-ablation}.

We also see that using fewer anchor trajectories for motion estimation leads to more average displacement, but with strong losses in rigidity, isometry and also appearance, indicating a stronger scattering of the 3D Gaussians.
Using more anchor trajectories for the deformation estimation leads to opposite behavior, where motion is more rigid, but on average less movement is generated due to the smoothing effect of using too many anchors.

\mysection{Limitations}
As outlined in \cref{sec:main-limitations}, our proposed method exhibits several limitations. In this section, we will provide more details on these shortcomings.

\begin{figure}
    \centering
    \includegraphics[width=\columnwidth]{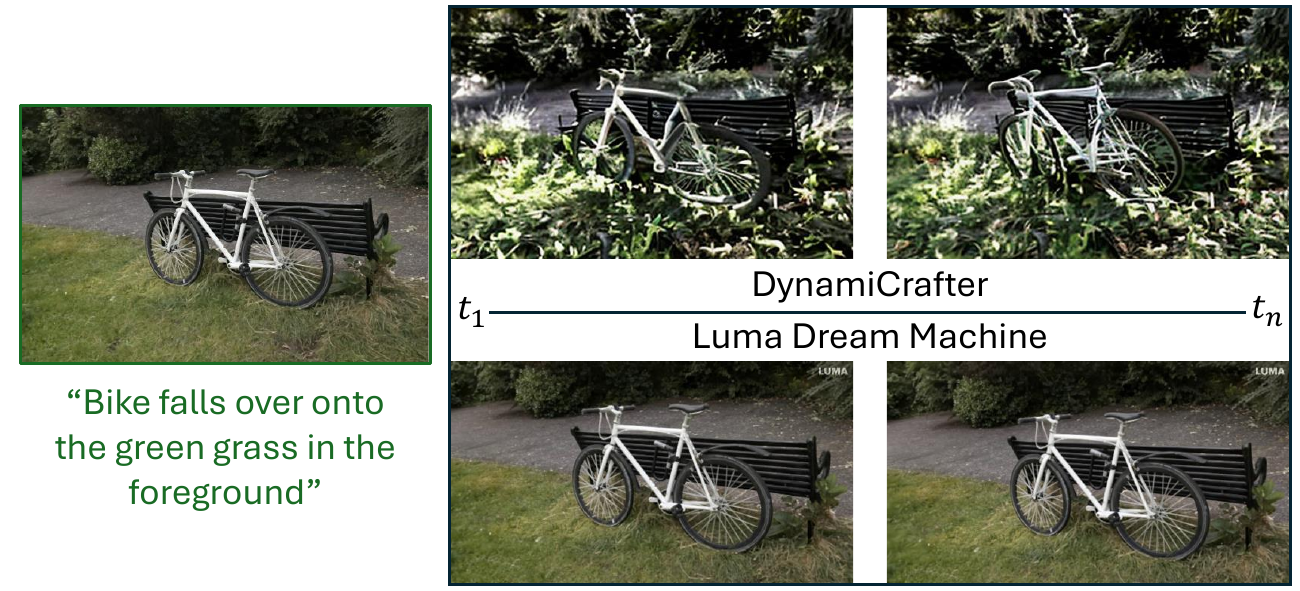}
    \caption{Failed generations of the video diffusion model. While open video diffusion models like DynamiCrafter~\citep{xing2023dynamicrafter} sometimes result in generation of artifacts when queried with OOD-samples, even more closed-source models like \href{https://lumalabs.ai/dream-machine}{Luma Dream Machine} struggle with realistic generation of the desired motion.}
    \label{fig:video-diffusion-failures}
\end{figure}

\begin{figure*}[htb]
    \centering
    \includegraphics[width=\textwidth]{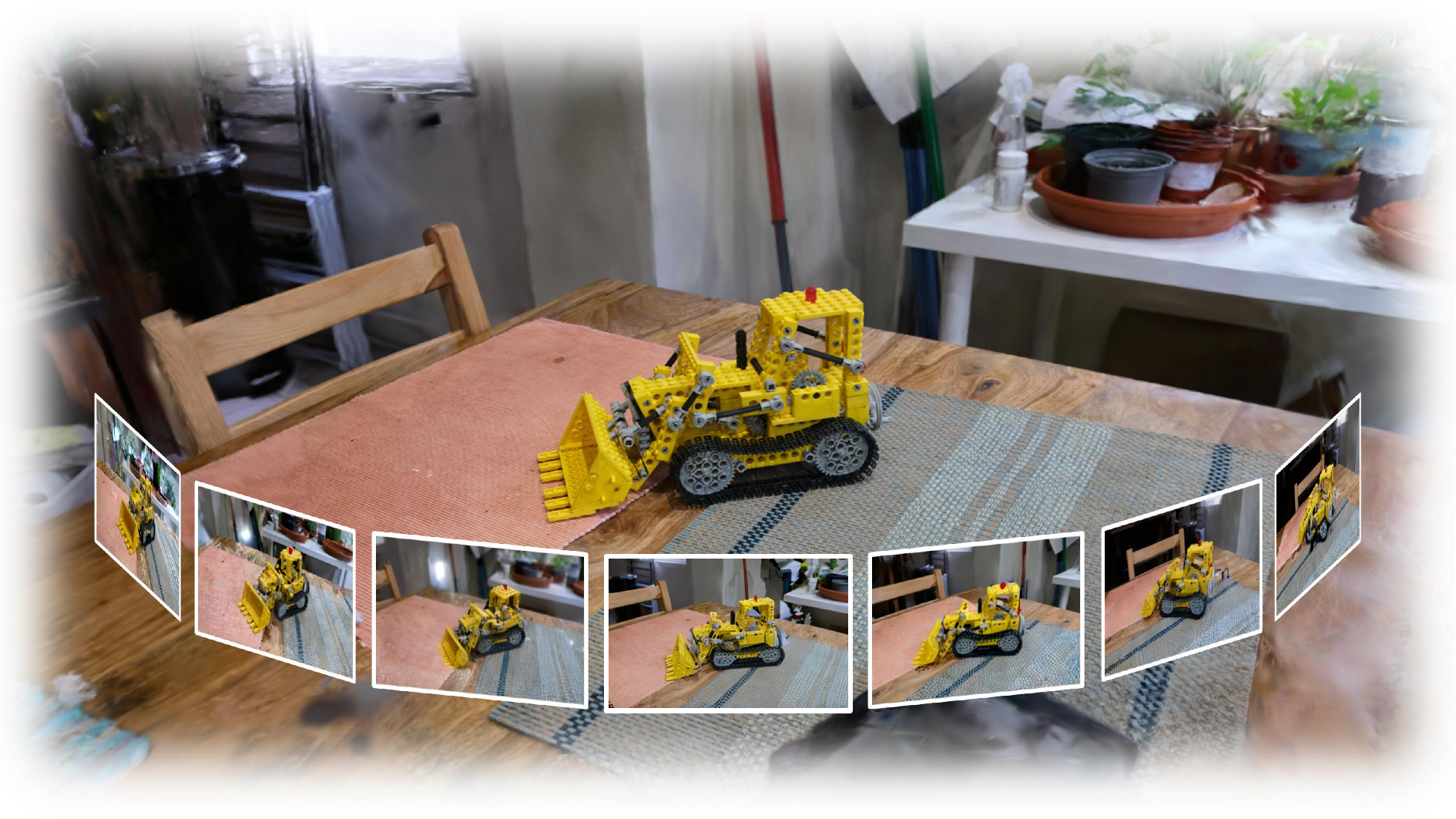}
    \caption{Approximately multi-view consistent video generations, starting from one video generated from an anchor viewpoint. The 3D scene is kept static between all generation steps to demonstrate the effect of the proposed latent interpolation. See the supplementary videos that also contain examples without the proposed latent interpolation for a comparison.\label{fig:mv-diffusion-example}}
\end{figure*}

While we are able to perform faithful deformations of given 3D scenes, our method is currently mainly limited by the guiding video diffusion model. More specifically, current \textit{open} video diffusion models lack camera control and are often inconsistent in their generations, both for the same view with different random seeds and for multi-view generation with the same seed. As main limitation, we found that the text conditioning often does not succeed  and inappropriate motion is generated. Additionally, resulting videos can exhibit artifacts, or heavily diverge from the given image condition to resemble more in-distribution data. In \cref{fig:video-diffusion-failures}, we show a examples for these video diffusion model failures.

Another limitation of our method is that it is currently unable to compensate for camera motion in the guidance videos. While prompting (e.g.,~appending ``static camera''), or using explicit camera-posed video diffusion models can help alleviate this problem, a more sophisticated solution would be to estimate camera poses before projecting motion from 2D videos to 3D, using similar techniques as, e.g., proposed in monocular dynamic reconstruction works~\citep{lei2024mosca}.

Finally, as mentioned in \cref{sec:main-limitations}, the proposed method only deforms given 3D scenes without adding or removing Gaussians. This can result in the emergence of ``holes'', as can be seen in \cref{fig:qualitative-results} for the toy bulldozer scenes. Another limitation that can be observed in these scenes is the reliance on object masks for deformation, as otherwise neighboring Gaussians can be easily deformed together with the object itself. This is caused by the point cloud nature of the 3DGS representation, where there is no clear notion on which Gaussians belong to the same object. However, as mentioned before, obtaining these masks is possible using off-the-shelf open-world 3D segmentation models~\citep{qin2023langsplat}.

\mysection{Ethical Considerations}
Finally, we would like to briefly address some ethical considerations in connection with our method. While the current results do not yet harbor potential dangers due to the brevity of the generated motion, it should not go unmentioned that the presented method, like all generative models, harbors the danger of deception, e.g., by means of deep fakes. While there are specialized methods for animating people and faces~\citep{shao2024splattingavatar, abdal2023gaussian, pang2023ash, jung2023deformable} that could be more dangerous in these aspects, our method is still not without danger, as it can be applied to all kinds of objects and scenes. It is therefore necessary that such methods are developed and used responsibly, with clear guidelines and oversight to prevent misuse.

\end{document}